\documentclass[10pt,twocolumn,letterpaper]{article}

\usepackage[pagenumbers]{wacv} 

\usepackage{graphicx}
\usepackage{amsmath}
\usepackage{amssymb}
\usepackage{booktabs}
\usepackage{times}
\usepackage{microtype}
\usepackage{epsfig}
\usepackage{caption}
\usepackage{float}
\usepackage{placeins}
\usepackage{color, colortbl}
\usepackage{stfloats}
\usepackage{enumitem}
\usepackage{tabularx}
\usepackage{xstring}
\usepackage{multirow}
\usepackage{xspace}
\usepackage{url}
\usepackage{subcaption}
\usepackage{xcolor}
\usepackage{afterpage}
\usepackage{blindtext}
\usepackage[hang,flushmargin]{footmisc}
\usepackage{wentinn}

\usepackage{pifont}
\newcommand{\cmark}{\ding{51}}%
\newcommand{\xmark}{\ding{55}}%

\usepackage[pagebackref,breaklinks,colorlinks]{hyperref}

\usepackage[capitalize]{cleveref}
\crefname{section}{Sec.}{Secs.}
\Crefname{section}{Section}{Sections}
\Crefname{table}{Table}{Tables}
\crefname{table}{Tab.}{Tabs.}

\def\wacvPaperID{271} 
\def\confName{WACV}
\def\confYear{2026}

\begin{document}

\title{Artifacts and Attention Sinks: Structured Approximations \\ for Efficient Vision Transformers}


\author{
Andrew Lu, \, Wentinn Liao, \, Liuhui Wang, \, Huzheng Yang, \, Jianbo Shi \\
University of Pennsylvania \\
{\tt\small \{alu1, wenliao, wanglh, huze, jshi\}@seas.upenn.edu}
}

\maketitle

\begin{abstract} 
Vision transformers have emerged as a powerful tool across a wide range of applications, yet their inner workings remain only partially understood. In this work, we examine the phenomenon of massive tokens—tokens with exceptionally high activation norms that act as attention sinks—and artifact tokens that emerge as a byproduct during inference. Our analysis reveals that these tokens mutually suppress one another through the attention mechanism, playing a critical role in regulating information flow within the network. Leveraging these insights, we introduce Fast Nystr\"om Attention (FNA), a training-free method that approximates self-attention in linear time and space by exploiting the structured patterns formed by massive and artifact tokens. Additionally, we propose a masking strategy to mitigate noise from these tokens, yielding modest performance gains at virtually no cost. We evaluate our approach on popular pretrained vision backbones and demonstrate competitive performance on retrieval, classification, segmentation, and visual question answering (VQA), all while reducing computational overhead. \end{abstract}

\section{Introduction}
\label{sec:intro}

Vision transformers have rapidly become a cornerstone in modern computer vision, achieving state-of-the-art results in tasks ranging from image classification to object detection and segmentation \cite{he2022masked} \cite{oquab2023dinov2} \cite{radford2021clip} \cite{touvron2022deitiii}. These models leverage the transformer architecture to process images as sequences of patches. With their ability to model long-range dependencies and capture global context, vision transformers \cite{dosovitskiy2021vit} have demonstrated remarkable performance across a wide variety of benchmarks \cite{chen2015coco} \cite{deng2009imagenet} \cite{pascal-voc-2012} \cite{young2014flickr} \cite{zhou2017ade}.

The unique designs of vision transformers have given rise to intriguing behaviors that are not yet fully understood. One such phenomenon is the emergence of a subset of tokens that exhibit exceptionally high activation norms in certain layers of the network. These tokens, which we refer to as ``\textit{massive tokens}'' (occasionally abbreviated as ``MA''), appear to dominate the attention distribution, effectively acting as ``\textit{attention sinks}'' \cite{gu2024attnsink} that influence the overall flow of information through the network.

In addition to massive tokens, our investigations reveal the presence of what we term ``\textit{artifact tokens}.'' These tokens do not naturally exhibit the extreme activation norms of massive tokens; however, they become evident under specific conditions---taking on the extreme-magnitude and attention-sink characterization of massive tokens only when the original massive tokens have been masked or removed. This observation suggests that vision transformers possess a built-in redundancy mechanism, where a limited number of tokens are capable of assuming the role of massive tokens if needed. 

These observations carry both theoretical interest and practical implications. We demonstrate that by strategically leveraging the distinct roles of massive and artifact tokens, it is possible to reconfigure the model’s attention dynamics to improve computational efficiency and boost performance.

In light of these observations, our work primarily makes the following contributions:
\begin{itemize}
    \item Fast Nystr\"om Attention (FNA), a training-free method that approximates self-attention at inference in linear time and space complexity using the properties of massive and artifact tokens.
    \vspace{-0.2cm}
    \item A novel, training-free algorithm that efficiently identifies massive tokens in vision transformers, enabling extraction and strategic masking with minimal computational overhead during inference. We show that our proposed masking procedure yields consistent performance improvements across a range of downstream multi-modal and dense prediction tasks.
    \vspace{-0.2cm}
    \item A comprehensive analysis of the mechanisms in vision transformers that facilitate the formation of massive tokens and their correlation with artifact tokens.
\end{itemize}

\begin{figure*}[t]
    \centering
    \includegraphics[width=1\linewidth]{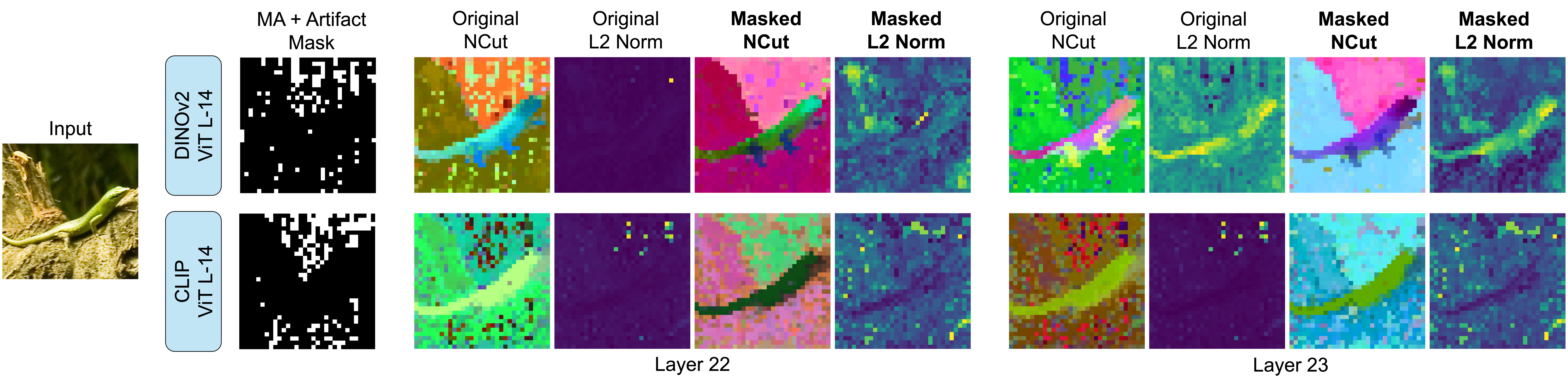}
    \vspace{-2em}
    \caption{Visualizations of sink (MA + Artifact) tokens masks applied to image features in the last two layers of CLIP~\cite{radford2021clip} and DINOv2~\cite{oquab2023dinov2}. Masks are constructed from our iterative detection method described in \ref{sec:iterative_masking}. Feature visualizations performed with NCut \cite{yang2024ncut} and L2 normalization show both the visual appearance of sink tokens and that masking can denoise features, emphasizing regions of interest with no additional training needed. More visualizations can be found in Appendix \ref{sec:appendix_zoo}.}
    \label{fig:dino_clip_artifacts}
\end{figure*}

\section{Related Work}
\label{sec:related}

\subsection{Massive and Artifact Tokens}

Massive tokens in transformer models have been recognized as an important phenomenon that heavily influences model behavior. Previous research \cite{darcet2024vision} \cite{gu2024attnsink} \cite{sun2024massive} \cite{yu2024super} has identified these tokens as constituting a disproportionate amount of attention in the middle to late layers of large pretrained transformers, effectively acting as attention sinks. Notably, Sun et al. demonstrated that the presence of massive tokens is vital to overall model performance in language models~\cite{sun2024massive}. 

Other studies \cite{darcet2024vision} \cite{yang2023emernerf} \cite{yang2024denoising} have noted the appearance of noisy artifacts in the intermediate and output features of self-supervised vision transformers such as CLIP \cite{radford2021clip} and DINOv2 \cite{oquab2023dinov2}. Specifically, Yang et al. proposed training a secondary denoising network to remove these artifacts, enjoying a performance gain on downstream tasks as a result ~\cite{yang2024denoising}. Darcet et al. \cite{darcet2024vision} observed that the quantity of massive tokens can be reduced by introducing register tokens in the training process; however, this does not fully resolve the emergence of artifacts \cite{yang2024denoising}. 

Despite the clear importance of massive tokens for overall model performance, little work has been dedicated to analyzing the underlying mechanisms of their formation. Even fewer studies have examined the landscape of artifacts, leaving significant opportunity for further research in this area.

\subsection{Efficient Attention}
The quadratic computational and memory requirements of the self-attention mechanism \cite{vaswani2017attention} in transformers have led to the development of various approaches to reduce its cost. Sparse attention techniques \cite{chen2023sparse} \cite{child2019sparseattn} \cite{ren2021combiner} limit the number of dot-product operations by only attending to a subset of tokens, while models such as the Linformer \cite{wang2020linformer} and Longformer \cite{beltagy2020longformer} employ structured sparse patterns (e.g., local windowed attention with task-specific global tokens) to achieve linear or near-linear complexity. These methods, along with others like Performer \cite{choromanski2020performer} that leverage random feature approximations, have shown promising improvements in scaling self-attention to longer sequences. 

In particular, the Nystr\"om-based approach proposed by Xiong et al. \cite{xiong2021nystromformer} approximates the full attention matrix by sampling a subset of its columns and rows, thereby reducing the quadratic complexity to a function of the number of samples. However, many of these methods require additional training or finetuning to achieve state-of-the-art performance, highlighting an open challenge to develop training-free alternatives.

\section{Massive and Artifact Tokens}
\label{sec:artifact_tokens}

\subsection{Notation}
While vision transformers may vary in architecture, most such as CLIP and DINO employ the one proposed by \cite{dosovitskiy2021vit}. We represent a tokenized image as a sequence of $N$ input embeddings $\emb{0}{1}, \emb{0}{2}, \ldots, \emb{0}{N} \in \R^D$ along with the class (CLS) token $\emb{0}{CLS}$, which are vertically stacked to compose $\Emb{0} = \tr{[\emb{0}{CLS} \quad \emb{0}{1} \quad \cdots \quad \emb{0}{N}]} \in \R^{(N + 1) \times D}$. For ease of notation, $\emb{\ell}{0}$ will be equivalent to $\emb{\ell}{CLS}$. Throughout the paper, layers, both as part of equations and explicitly referred to, will be zero-indexed. We define an $L$-layer transformer to be a sequence $(\lyr{0}, \ldots, \lyr{L - 1})$ where $\lyr{\ell}$ is equipped with the $4$-tuple of functions $(\lyn{1}{\ell}, \attn{\ell}, \lyn{2}{\ell}, \mlp{\ell})$, and \begin{alignat}{3}
    & \Emb{\ell + 1/2}  && = \Emb{\ell} + \attn{\ell}(\lyn{1}{\ell}(\Emb{\ell})), \label{eqn:layer}   \\
    & \Emb{\ell + 1}    && = \Emb{\ell + 1/2} + \mlp{\ell}(\lyn{2}{\ell}(\Emb{\ell + 1/2})), \\
    & \Emb{\ell + 1}    && = \lyr{\ell}(\Emb{\ell}).
\end{alignat} Although each of these four functions are parameterized by some set of learned weights, our analysis is primarily concerned with those composing the attention operation. In a multi-head attention operation with $H$ distinct heads, these weights are denoted as $\cbn{(\Ql{\ell}_h, \Kl{\ell}_h, \Vl{\ell}_h, \Olw{\ell}{h})}_{h \in [H]}$ and $\Olb{\ell}$ where $H \cdot d = D$, $\Ql{\ell}_h, \Kl{\ell}_h, \Vl{\ell}_h: \R^D \to \R^{d}$, $\Olw{\ell}{h} \in \R^{D \times d}$, and $\Olb{\ell} \in \R^D$. For the sake of brevity, $\Ql{\ell}_h(\lyn{1}{\ell}(\Emb{\ell}))$ can be abbreviated as $\Ql{\ell}_h \in \R^{(N + 1) \times d}$, with $\Kl{\ell}_h$ and $\Vl{\ell}_h$ being abbreviated identically. Where $\Embp{\ell} = \lyn{1}{\ell}(\Emb{\ell})$, the attention operation is given by \begin{align}
    \attn{\ell}(\Embp{\ell}) = \Olb{\ell} + \dss{h}{} \sf\pfrac{\Ql{\ell}_h{\mbf{K}^{(\ell)\top}_h}}{\sqrt{d}}\Vl{\ell}_h{\mbf{O}^{(\ell)\top}_{h, weight}}, \label{eqn:attn}
\end{align} where $\sf$ denotes the softmax operator, which will only refer to its application on the last dimension to avoid ambiguity in cases where its operand tensor has more than one dimension. For an attention matrix $\Al{\ell}_h = \sf(\Ql{\ell}_h{\mbf{K}^{(\ell)\top}_h} / \sqrt{d})$, we primarily employ the indexing notation $\Al{\ell}_{h, i \to j}$ to emphasize that that entry represents the attention from $i$ to $j$.

\subsection{Attention Sinks}
\label{sec:attention_sinks}
\begin{figure*}[t]
    \centering
    \includegraphics[width=1\linewidth]{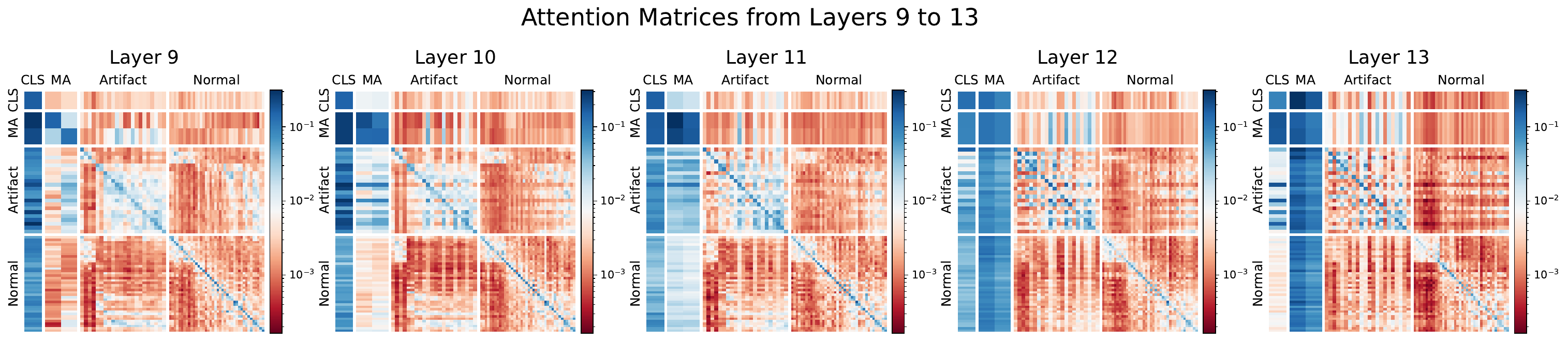}
    \vspace{-1.5em}
    \caption{Each subplot visualizes the mean attention matrix across heads for a single image at a particular layer of CLIP ViT L-14. In an image tokenization with $256$ image tokens and one CLS token, the mean attention matrix is $257 \times 257$ which is difficult to visualize. We permute the order of tokens and resize sections of the attention matrix to distinguish small subsets of tokens that are of interest, particularly massive tokens, and artifact tokens. Additionally, we subsample a portion of the remaining tokens which allows us to see them in detail as well as eliminate any perceptive bias resulting from the discrepancy in scale. We can see that in layers 9 and 10, massive tokens are characterized by large attention to the CLS token and large attention to itself, and become attention sinks in layers 11 to 13 where they attract a large proportion of attention from all tokens.}
    \label{fig:attention_transition_layers}
\end{figure*}

While massive tokens are most easily observed through their large activation norms, they also attract a large proportion of attention from all tokens. We find that the constitution of a large proportion of attention within very few tokens is critical to generating effective feature representations; however, their presence in later layers modestly detracts accumulation of information into the class (CLS) token. As seen in Figure \ref{fig:dino_clip_artifacts}, the denoising of such tokens improves the quality and coherency of feature representations in the final layers.

We also observe that the incoming attention to massive tokens dramatically increases over their formative layers as seen in Figure \ref{fig:attention_transition_layers}. Though the massive tokens that emerge naturally from a model number very few (approximately 2-3 per image), we find that models actually learn a robust process that enables other suitable tokens to become massive in their place should the original massive tokens be removed. Furthermore, those tokens are roughly ordered, in which a suitable token will become massive only if a sufficient set of tokens that precede it have been removed via masking. We denote these dormant tokens as \textit{artifact tokens}, and the pool of (potential) massive tokens as a whole as \textit{attention sinks}. Any image (non-CLS) token that is neither a massive token nor artifact token is referred to as a \textit{normal token}.

Since attention sinks serve a critical role in attracting attention, extracting such tokens can be intuitively performed by thresholding their average incoming attention after formation. However, we also find that attention \textit{from} the CLS token provides a more distinct signal for determining attention sinks. Specifically, we select a detection layer $\ld$ such that if $\Al{\ld} \in \R^{(N + 1) \times (N + 1)}$ represents the mean attention matrix in layer $\ld$, then token $t$ is labeled an attention sink if $\Al{\ld}_{CLS \to t} \geq \Al{\ld}_{CLS \to CLS}$, i.e. if the mean attention from CLS to $t$ exceeds the mean attention from CLS to itself.

This detection layer is typically $13$ for CLIP and $20$ for DINOv2 ViT L-14. While not all massive tokens meet the CLS threshold, it is also true that not all potential massive tokens exhibit immediate largeness or attention-sinkness. Therefore,  this observation alone does not suggest a method that is able to determine all such tokens within one iteration. In Section \ref{sec:iterative_masking}, we discuss an iterative procedure based on this intuition for determining both the set of attention sink tokens and their priority order. Additionally, in Section \ref{sec:artifact_classifier}, we present a fast, non-iterative method that is able to determine the set of attention sink tokens alone.



\subsection{Iterative Detection of Attention Sinks}

\label{sec:iterative_masking}
Our iterative procedure is formalized to extract all (potential) attention sinks. While the pseudocode (Algorithm \ref{alg:iterative_sink_detection}) is provided in concrete detail pertaining to CLIP ViT L-14, an analogous method can be applied to other vision transformers.
\begin{algorithm}[H]
    \caption{Attention Sink Detection via Iterative Removal}
    \hspace*{\algorithmicindent} \textbf{Input:} $\Emb{0} \in \R^{(N + 1) \times D}$ \\
    \hspace*{\algorithmicindent} \textbf{Output:} List $\mc{T} = (t_1, t_2, \ldots) \subseteq [N]$ of attention sinks.
    \begin{algorithmic}[1]
        \Procedure{ComputeSinks}{$\Emb{0}$}
            \State $\lm, \ld \gets 9, 13$
            \State Run transformer layers $0$ to $\lm - 1$
            \State Initialize $\mc{T}$ as $\emptyset$
            \While{$\mc{T}$ has not converged}
                \State Rerun $\lm$ to $\ld$ masking $\mc{T}$
                \State Let $A$ be the mean attention matrix at $\ld$
                \State Append $t$ to $\mc{T}$ if $A_{CLS \to t} > A_{CLS \to CLS}$
            \EndWhile
            \State \Return $\mc{T}$
        \EndProcedure
    \end{algorithmic}
    \label{alg:iterative_sink_detection}
\end{algorithm}
\vspace{-8pt}

\begin{figure*}[t]
    \centering
    \includegraphics[width=1\linewidth]{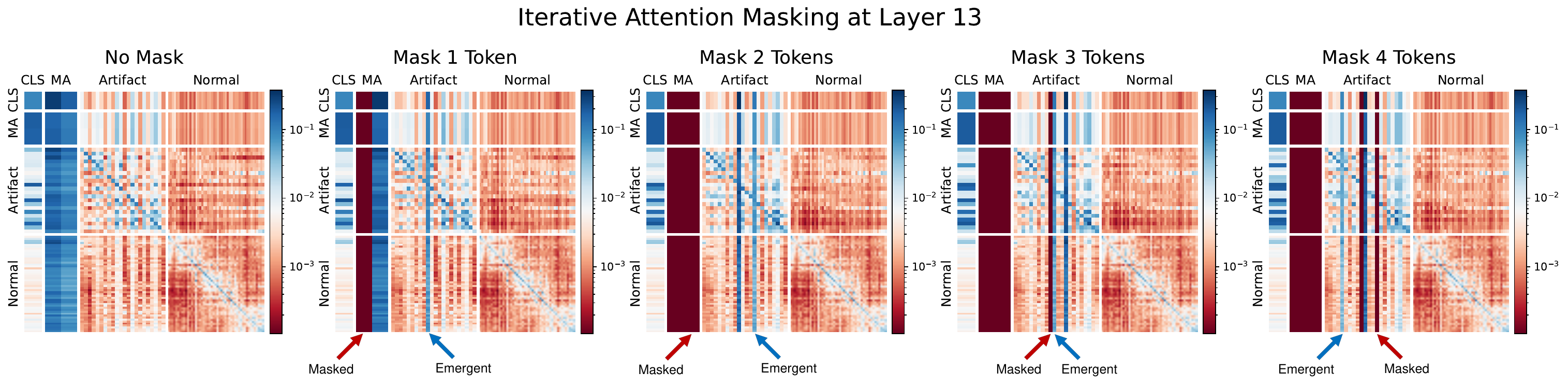}
    \vspace{-2em}
    \caption{Each subplot depicts the result of a step in the iterative removal process for a single image in CLIP ViT L-14. The leftmost attention matrix shows that massive tokens emerge as the natural attention sinks. However, each subsequent step masks out the most prominent attention sink (indicted with the red arrow) which results in the emergence of a new attention sink as a substitute (indicated with the blue arrow).}
    \label{fig:iterative_removal}
\end{figure*}

As shown in Figure \ref{fig:iterative_removal}, the masking of tokens results in the gradual emergence of substitute tokens that become both massive and attention sinks in their place.

\subsection{Non-iterative Detection of Attention Sinks}
\label{sec:artifact_classifier}

While our iterative method for extracting massive tokens is both precise and interpretable, it incurs a significant computational cost due to the need for multiple passes through the model's intermediate layers. However, in practice, massive tokens become readily identifiable shortly after their formation, as evidenced by feature visualization techniques (see Figure \ref{fig:dino_clip_artifacts}). Consequently, we can approximate the results of our iterative algorithm using traditional discrete clustering methods—such as Multi-class Spectral Clustering \cite{shi2000ncut} \cite{yang2024ncut}—to achieve a more computationally efficient, non-iterative classification of massive and artifact tokens.

\section{Fast Nystr\"om Attention}
\label{sec:fna}

The emergence of massive tokens as attention sinks creates a highly structured and predictable pattern in the attention matrix, particularly in the middle-to-late layers of vision transformers. Once these tokens form, they dominate the attention distribution, creating a low-rank structure where most queries primarily attend to their immediate context and sink tokens(see Figure \ref{fig:attention_transition_layers}). This phenomenon suggests that the full attention matrix—while quadratic in size—can be efficiently approximated by preserving the critical interactions involving these key tokens while compressing less informative regions. 

\subsection{Formulation}
\label{sec:formulation}

By leveraging token partitioning information, we can compress the attention matrix to achieve significant memory and speed improvements during inference. The primary objective is to construct a low-rank approximation of the attention matrix $\Ll{\ell}_hR_h^{(\ell)\top} \approx \Al{\ell}_h = \sf(\Ql{\ell}_h \mbf{K}^{(\ell)\top}_h / \sqrt{d})$ where $\Ll{\ell}_h, \Rl{\ell}_h \in \R^{(N + 1) \times s}$ for $s \ll N + 1$, which allows us to approximate the attention in $O(sND)$ time and space rather than $O(N^2D)$. The classical Nystr\"om extension suggests the approximation
\vspace{-4pt} \begin{align}
    \Al{\ell}_h
    \approx \Al{\ell}_{h, : \to S} (\Al{\ell}_{h, S \to S})^{-1} \Al{\ell}_{h, S \to :}
\end{align} where $S \subseteq [N]_0, \abs{S} = s$ represents the set of sampled tokens used for the quadrature approximation and $[N]_0$ represents the set of all integers from $0$ to $N$. However, we observe that exactly computing $\Al{\ell}_{h, : \to S}$ does not avoid quadratic complexity, as it would necessitate computation of all exponential row sums for the denominators of softmax. \cite{xiong2021nystromformer} resolves this issue by instead approximating with \vspace{-4pt} \begin{align}
    \resizebox{\columnwidth}{!}{$
        \Al{\ell}_h
        \approx \sf\pfrac{\Ql{\ell}_h k^{(\ell)\top}_h}{\sqrt{d}} \sf\pfrac{\ql{\ell}_h k^{(\ell)\top}_h}{\sqrt{d}}^{-1} \sf\pfrac{\ql{\ell}_h \mbf{K}^{(\ell)\top}_h}{\sqrt{d}}
    $}
\end{align}
where $\ql{\ell}_h, \kl{\ell}_h \in \R^{s \times d}$ are ``landmark features'' that approximate the set of tokens, opting to apply softmax on the intermediate matrices in spite of the discrepancy in softmax denominator.

\subsection{Methods}

While our method of decomposition is drawn from \cite{xiong2021nystromformer}, our work differs in two key ways: 
\begin{enumerate}[label=\arabic*)]
    \item we opt for a universally-applicable training-free approach that aims to improve the efficiency of vision transformers without the need to modify the underlying model, and
    \vspace{-0.2cm}
    \item while \cite{xiong2021nystromformer} uses Segment-Means cluster centers as their landmark features, we instead choose to sample points directly from the set of tokens via Farthest Point Sampling (FPS) \cite{qi2017fps}, producing better results in comparison to Segment-Means and other training-free approaches, and bypassing the need to compute cluster centers at inference time.
\end{enumerate}

We identify CLS, massive tokens, and artifact tokens as three sets of tokens that we may wish to regard differently from the remainder of the tokens while sampling. Let $\fpsl{\ell}(S, k) = F$ denote the procedure in which we sample $k$ points from the block outputs $\cbn{\emb{\ell}{i}}_{i \in S}$ via farthest point sampling and return their indices so that $F \subseteq S$, $\abs{F} = k$. Then, given a ``guarantee'' set $G$ and an ``exclusion'' set $E$ where $G, E \subseteq [N]_0$ and $G \cap E = \emptyset$, we sample $s$ points from $\cbn{\emb{\ell}{CLS}, \emb{\ell}{0}, \ldots, \emb{\ell}{N}}$ by \begin{align}
    S = G \cup \fpsl{\ell}([N]_0 \setminus (G \cup E), s - \abs{G}).
\end{align} 
To simplify, we sample $s$ points with the guarantees that all points in $G$ are sampled, no point in $E$ is sampled, and the sample quota not covered by $\abs{G}$ is sampled from the remaining points via FPS. We can therefore regard each of the interest sets in three ways: to assign them to $G$ (``guarantee''), assign them to $E$ (``exclude''), or assign them to the FPS sampling pool $[N]_0 \setminus (G \cup E)$ (``ignore''), which results in a total of $3^3 = 27$ different configurations.


In our Fast Nystr\"om Attention method, we guarantee the inclusion of the CLS token while using FPS to select the remaining tokens. This approach works effectively because massive and artifacts tokens are statistical outliers on the feature manifold, ensuring that FPS naturally represents the sink token population without over-saturating the subsample. In comparison to other sampling methods, we find that this performs nearly identical to guaranteeing the sampling of massive tokens, significantly better than guaranteeing the sampling of artifact tokens, and notably better than excluding either. We opt to ignore rather than guarantee massive tokens as it bypasses the need to explicitly extract them and instead relegate that task to the proficiencies of FPS. The details of these results are illustrated in Appendix \ref{sec:appendix_results_fna_sampling}.

\begin{table}[H]
    \centering
    \footnotesize
    \resizebox{\columnwidth}{!}{%
        \begin{tabular}{lcccc}
            \toprule
             & \multicolumn{2}{c}{Standard Attention} & \multicolumn{2}{c}{Fast Nystro\"m Attention \textbf{(ours)}} \\
            Sequence Length & Memory (MB) & Time (ms) & Memory (MB) & Time (ms) \\
            \midrule
            256 & 133 & 0.8 & 108 & 1.2 \\
            512 & 257 & 2.1 & 140 & 2.0 \\
            1,024 & 697 & 5.4 & 204 & 3.4 \\
            2,048 & 2,376 & 17.5 & 332 & 6.0 \\
            4,096 & 8,904 & 55.7 & 588 & 11.6 \\
            8,192 & 34,632 & 201.8 & 1,100 & 22.9 \\
            \bottomrule
        \end{tabular}%
    }
    \vspace{-0.5em}
    \caption{Memory consumption and running time results on various sequence lengths. We report the average memory consumption and running time for one input batch (batch size = 8) through a standard self-attention module (from scratch) and our Fast Nystr\"om Attention (sample size = 64). }
    \label{tab:nfa_speed_and_mem}
\end{table}


The main bottleneck of Fast Nystr\"om Attention lies in the FPS sampling step which is necessary to reduce the time complexity of the attention mechanism for each block from $O(N^2D)$ to $O(sND)$ and its space complexity to from $O(N^2 + ND)$ to $O(sN + ND)$. While the FPS subroutine itself requires $O(N^2D)$ time and $O(N^2)$ space to compute the pairwise distance matrix, we find that we can produce comparable results by sampling once after massive token formation and reusing those samples in the subsequent layers. This results in an overall reduction from $O(LN^2D)$ to $O(N^2D + sLND)$ in time, and while the peak memory consumption remains $O(N^2 + ND)$, it is reduced to $O(sN + ND)$ after Fast Nystr\"om Attention is applied. We compare the inference time and memory consumption of Fast Nystr\"om Attention with standard attention in Table \ref{tab:nfa_speed_and_mem}.

\section{Experiments}

We implemented Fast Nystr\"om Attention as a PyTorch module that serves as a drop-in replacement for standard attention. We evaluated our approach on CLIP \cite{ilharco2021openclip} \cite{radford2021clip} and DINOv2 \cite{oquab2023dinov2} ViT L-14 models without any additional training or fine-tuning. Retrieval performance was assessed on COCO Captions \cite{chen2015coco} and Flickr30k \cite{young2014flickr} datasets using Recall@K metrics for bidirectional text-image retrieval. For vision-specific applications, we conducted zero-shot classification on ImageNet \cite{deng2009imagenet} and linear probing for semantic segmentation on VOC2012 \cite{pascal-voc-2012} and ADE20k \cite{zhou2017ade} datasets. All experiments were performed using a single NVIDIA RTX 4090 GPU.

\subsection{Pretrained Vision Backbones}


Tables \ref{tab:fna_retrieval} and \ref{tab:fna_vision_only_results} summarize the results of applying Fast Nystr\"om Attention to CLIP and DINOv2. Our experiments show that Nystr\"om attention compression with FPS sampling delivers comparable results to standard attention on bidirectional retrieval, classification, and segmentation across multiple datasets. Notably, our one-time sampling strategy remains competitive with resampling at each layer and allows us to improve efficiency with only a minimal impact on performance metrics.

As seen in Table \ref{tab:fna_sampling_results}, Farthest Point Sampling runs only marginally slower than uniform random sampling while performing $~2\%$ better across all retrieval benchmarks, and outclasses other methods such as $k$-means and Spectral Clustering in both speed and performance. These results are supported by \cite{xiong2021nystromformer}, which shows that compression via Segment-Means requires retraining to achieve comparable results to standard attention. In addition, sampling methods that compute aggregate features (\textit{e.g.} Segment-Means) must be recomputed at every layer, unlike ``pure'' sampling methods.

There is an intuitive tradeoff between smaller sample sizes and model performance. Empirically, we find a sample size of 32 to 64 sufficient to approximate attention with competitive results to standard attention with no additional training (Appendix \ref{sec:appendix_results_fna_sampling}). Results of applying Fast Nystr\"om Attention to vision backbones of CLIP and DINOv2 are shown in Tables \ref{tab:fna_retrieval} and \ref{tab:fna_vision_only_results}.

\begin{table}[t]
    \centering
    \footnotesize
    \begin{subtable}[t]{\columnwidth}
        \centering
        \resizebox{\columnwidth}{!}{%
            \begin{tabular}{lcccccc}
                \toprule
                 & \multicolumn{3}{c}{COCO} & \multicolumn{3}{c}{Flickr30k} \\
                Model & R@1 & R@5 & R@10 & R@1 & R@5 & R@10 \\
                \midrule
                CLIP & 35.33 & 59.97 & 70.15 & 65.20 & 87.24 & 92.00  \\
                CLIP+FNA+no resample & 35.42 & 60.18 & 70.27 & 65.17 & 87.22 & 91.95 \\
                CLIP+FNA+resample & 35.58 & 60.43 & 70.51 & 65.27 & 87.25 & 91.98 \\
                \bottomrule
            \end{tabular}%
        }
        \caption{Image retrieval}
        \label{tab:fna_image_retrieval_sub}
    \end{subtable}
    
    \vspace{1em}
    
    \begin{subtable}[t]{\columnwidth}
        \centering
        \resizebox{\columnwidth}{!}{%
            \begin{tabular}{lcccccc}
                \toprule
                 & \multicolumn{3}{c}{COCO} & \multicolumn{3}{c}{Flickr30k} \\
                Model & R@1 & R@5 & R@10 & R@1 & R@5 & R@10 \\
                \midrule
                CLIP & 56.06 & 79.48 & 86.84 & 85.10 & 97.30 & 99.00 \\
                CLIP+FNA+no resample & 55.39 & 78.72 & 86.49 & 85.06 & 97.16 & 98.78 \\
                CLIP+FNA+resample & 55.60 & 79.04 & 86.65 & 85.13 & 97.29 & 98.99 \\
                \bottomrule
            \end{tabular}%
        }
        \caption{Text retrieval}
        \label{tab:fna_text_retrieval_sub}
    \end{subtable}
    \vspace{-1em}
    \caption{Zero-shot retrieval results for CLIP ViT L-14 on COCO \cite{chen2015coco} and Flickr30k \cite{young2014flickr}. Our Fast Nystr\"om Approximation (FNA) with a sample size of 64 achieves competitive results with standard CLIP with no finetuning necessary.}
    \label{tab:fna_retrieval}
\end{table}

\begin{table}[t]
    \centering
    \footnotesize
    \resizebox{\columnwidth}{!}{%
        \begin{tabular}{lcccccc}
            \toprule
             & \multicolumn{2}{c}{ImageNet} & \multicolumn{2}{c}{VOC2012} & \multicolumn{2}{c}{ADE20k} \\
            Model & Top1 & Top5 & aACC & mIoU & aACC & mIoU \\
            \midrule
            CLIP & 75.96 & 94.82 & 90.33 & 66.60 & 69.55 & 34.84 \\
            CLIP+FNA+no resample & 75.74 & 94.49 & 90.19 & 66.53 & 69.18 & 34.64 \\
            CLIP+FNA+resample & 75.81 & 94.82 & 90.24 & 66.57 & 69.25 & 34.67 \\
            \midrule
            DINOv2 & 78.62 & 92.91 & 94.12 & 77.54 & 78.65 & 44.48 \\
            DinoV2+FNA+no resample & 78.57 & 92.89 & 93.91 & 77.35 & 78.46 & 44.40 \\
            DinoV2+FNA+resample & 78.60 & 92.92 & 93.98 & 77.41 & 78.49 & 44.42 \\
            \bottomrule
        \end{tabular}%
    }
    \vspace{-0.5em}
    \caption{Classification and segmentation benchmarks for pretrained CLIP and DINOv2 ViT L-14 show that Our Fast Nystr\"om Approximation (FNA) with a sample size of 64 achieves competitive results on dense vision tasks with no finetuning. ImageNet \cite{deng2009imagenet} classification evaluation is performed in the zero-shot setting. Segmentation on VOC2012 \cite{pascal-voc-2012} and ADE20k \cite{zhou2017ade} is performed via fitting linear probes to output of the final layers.}
    \label{tab:fna_vision_only_results}
\end{table}


\begin{table}[t]
    \centering
    \footnotesize
    \resizebox{\columnwidth}{!}{%
        \begin{tabular}{lccccc}
            \toprule
             &  & \multicolumn{2}{c}{Text-to-Image} & \multicolumn{2}{c}{Image-to-Text} \\
            Model & Time (s) & R@1 & R@5 & R@1 & R@5 \\
            \midrule
            CLIP+FNA (Multiclass SC)    & 1572.12           & 34.22             & 59.09             & 54.04             & 77.90 \\
            CLIP+FNA (SC)               & 1563.95           & 35.36             & 59.99             & 55.68             & 79.06 \\
            CLIP+FNA (K-Means)          & 875.05            & 35.33             & 60.11             & 55.28             & \textbf{79.40} \\
            CLIP+FNA (Segment-Means) \cite{xiong2021nystromformer}    & 74.27             & 32.96             & 57.80             & 49.32             & 74.18 \\
            CLIP+FNA (Uniform)          & \textbf{56.63}    & 33.77             & 58.66             & 53.90             & 77.96 \\
            \midrule
            CLIP+FNA (FPS)              & 61.25             & \textbf{35.91}    & \textbf{60.43}    & \textbf{56.52}    & 79.32 \\
            \bottomrule
        \end{tabular}%
    }
    \vspace{-0.5em}
    \caption{Zero-shot retrieval time and performance results on COCO for CLIP ViT L-14 and its Fast Nyström Approximation (FNA) with different sampling strategies.}
    \label{tab:fna_sampling_results}
\end{table}



\subsection{Comparison with Existing Efficient Attention}
\begin{figure}[t]
    \centering
    \includegraphics[width=1\linewidth]{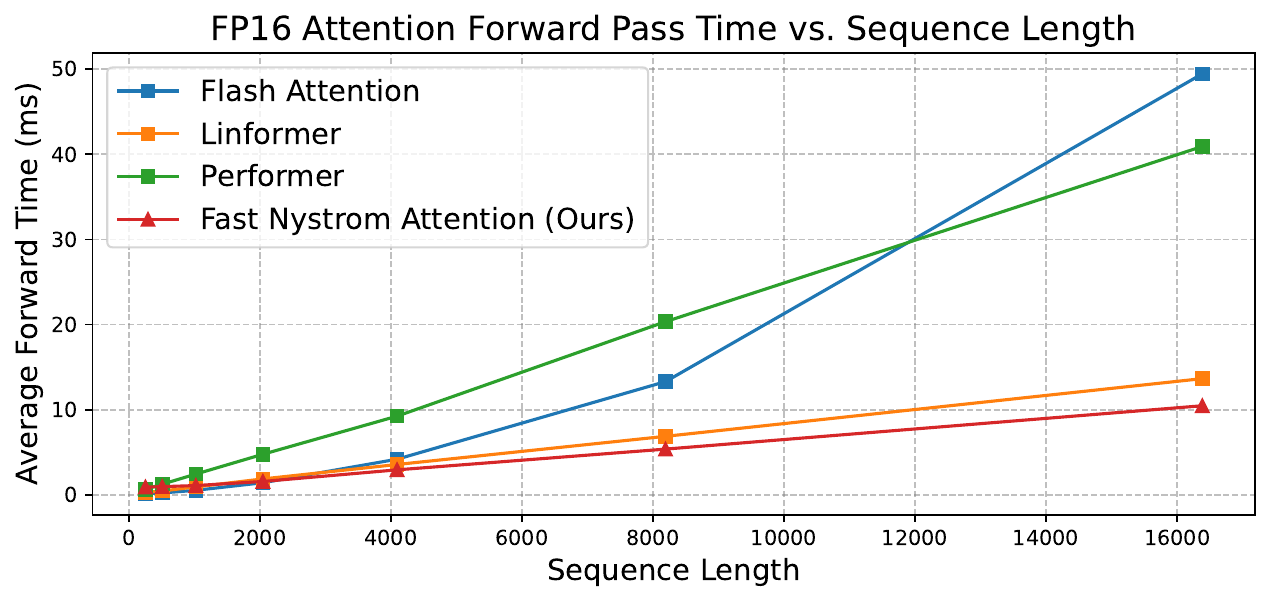}
    \vspace{-2em}
    \caption{Fast Nyström Attention outperforms existing linear attention methods \cite{choromanski2020performer} \cite{wang2020linformer} in inference speed and offers superior scaling to FlashAttention \cite{dao2024flashattention2}.}
    \label{fig:attention_timing}
\end{figure}

We benchmark Fast Nystr\"om Attention against existing efficient attention methods in Figure \ref{fig:attention_timing}, where it demonstrates highly competitive scaling. Compared to other linear scaling approximations \cite{choromanski2020performer} \cite{wang2020linformer}, our method is faster and training-free, allowing it to be applied as a drop-in replacement. While optimized exact attention like FlashAttention \cite{dao2024flashattention2} is popular, it remains fundamentally quadratic in time complexity. Results after finetuning are shown in Appendix \ref{sec:appendix_results_linear}.

\subsection{Vision Language Models}
\begin{figure}[t]
    \centering
    \includegraphics[width=1\linewidth]{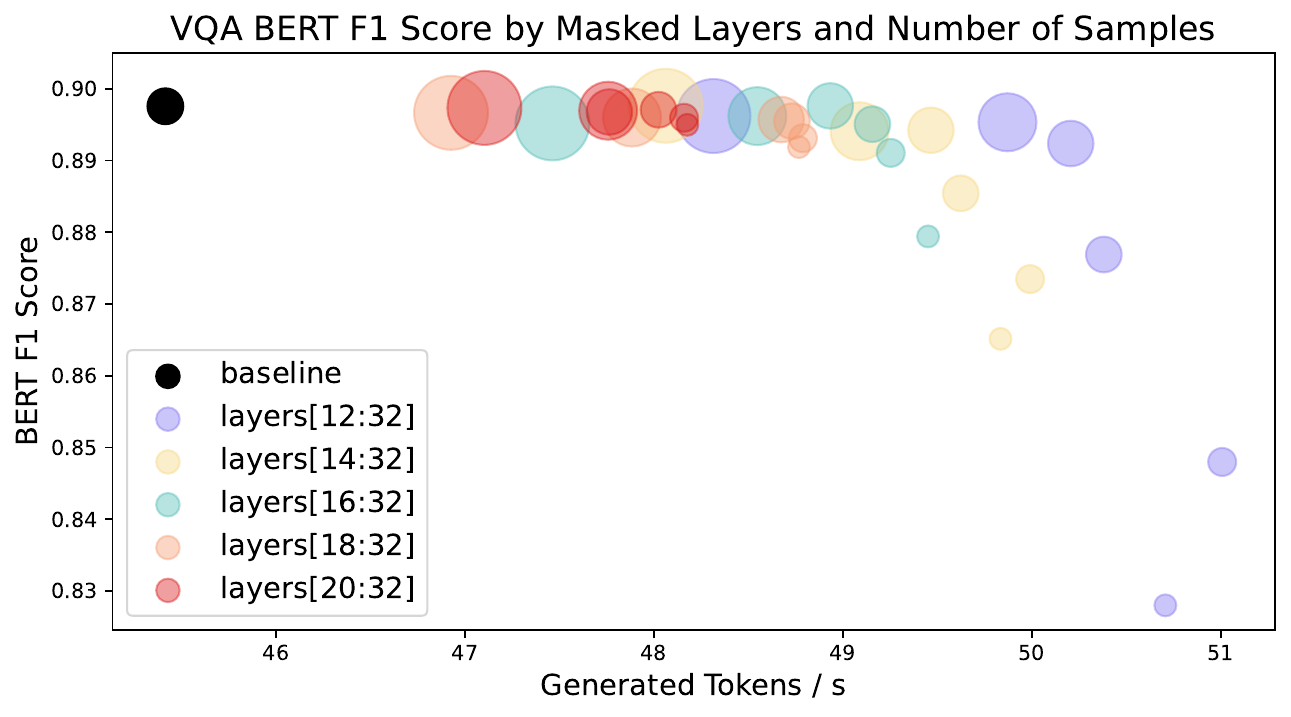}
    \vspace{-2em}
    \caption{BERTScore \cite{Zhang2020BERTScore} and generation speed on COCO VQA \cite{AgrawalCOCOVQA} for different configurations of Fast Nystr\"om Attention (FNA) applied to LLaVA-NeXT-7B \cite{liu2023llava}. Each color represents a LLaVA model with FNA applied on the causal attention to image tokens at the specified span of layers. Spot size represent FNA sample sizes ranging from 16 to 512 tokens out of $\sim2500$ image tokens. 
    }
    \label{fig:llava_bert_scores}
\end{figure}

We extend Fast Nyström Attention to LLaVA-NEXT-7B \cite{liu2023llava}, a vision-language model (VLM) composed of a pretrained image encoder and a large language model (LLM). In its standard operation, LLaVA processes an image into a large sequence of approximately 2500 tokens. These visual tokens are then incorporated into the LLM's causal attention mechanism to guide text generation, creating a significant computational load. While our reduction technique can be applied directly to LLaVA's vision encoder (as done for CLIP and DINOv2), reducing the tokens after they are projected into the LLM's text-embedding space proves more effective. Specifically, Nyström approximation is applied to the embedded image tokens within the LLaMA model, caching a compressed set of keys and values. This compact representation is then used for all subsequent causal attention steps, accelerating the generation of the text response. We evaluate on COCO VQA \cite{AgrawalCOCOVQA}, and report BERTScore \cite{Zhang2020BERTScore} between generated responses and ground-truth answers. As shown in Figure \ref{fig:llava_bert_scores}, Fast Nyström Attention boosts token throughput by 10\% while maintaining baseline performance. Qualitative examples are available in Appendix \ref{sec:appendix_results_llava}.

\begin{figure}
    \centering
    \includegraphics[width=1\linewidth]{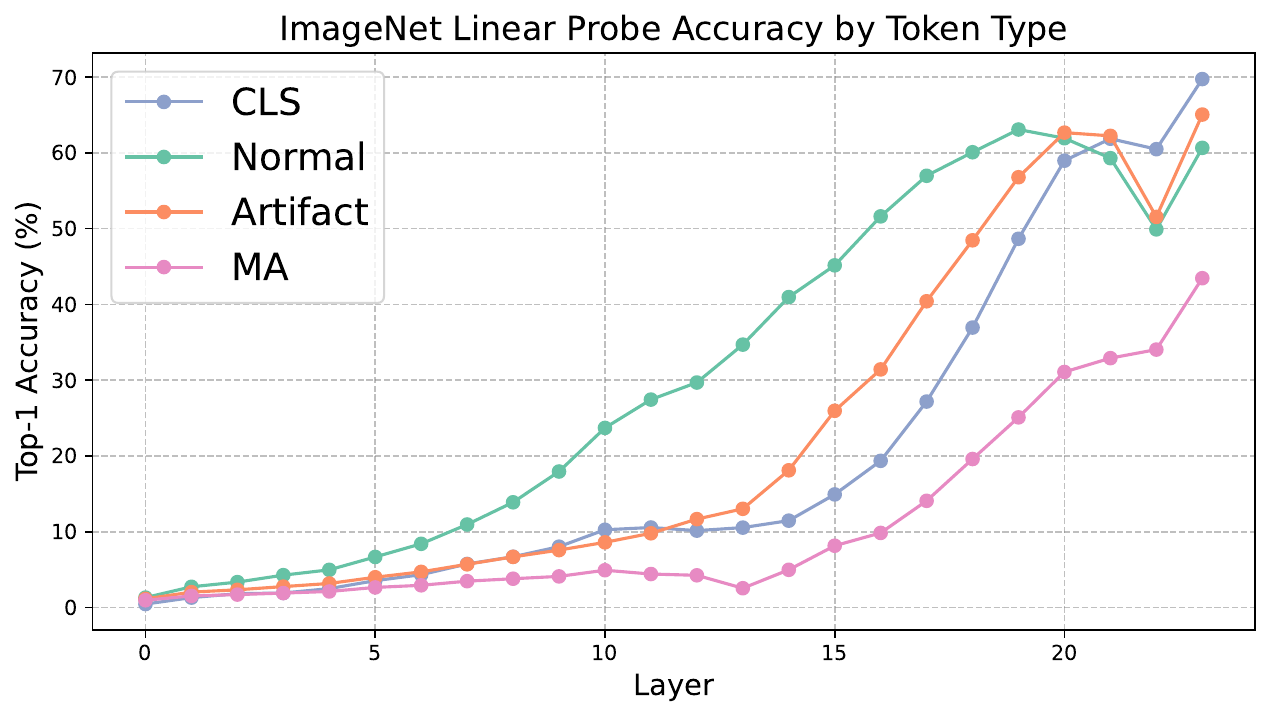}
    \vspace{-2em}
    \caption{Fitting a linear probe to the average token grouped by type at each layer shows that CLS tokens contain less global semantic information than Normal tokens until the last layers in CLIP.}
    \label{fig:imagenet_lin_probe}
\end{figure}

\section{Additional Performance Gains}
\label{sec:performance_efficiency}

Vision transformers that incorporate a CLS token often exhibit a subtle separation of information between the CLS and image tokens. To illustrate this, we evaluated a pretrained CLIP ViT L-14 model on ImageNet by fitting a linear probe to different token types at each layer. As shown in Figure \ref{fig:imagenet_lin_probe}, the CLS token initially contains less global semantic information than the average normal tokens as evidenced by lower scores in middle layers; however, it surpasses them in the final layers as it aggregates information for classification. Nevertheless, leftover massive and artifact tokens can interfere with the CLS token’s access to image features, effectively \emph{sinking} attention away. They also introduce noise into the patch representations themselves, degrading performance in both global (\eg, classification, retrieval) and dense tasks (\eg, segmentation).

In practice, we can detect massive and artifact tokens after formation in earlier layers and replace them with nearby normal tokens in the final layers. Applying this masking strategy to CLIP ViT L-14 consistently improves zero-shot retrieval on COCO Captions~\cite{chen2015coco} and Flickr30k~\cite{young2014flickr} (Table~\ref{tab:combined_retrieval}). A similar improvement on image classification and segmentation is shown in Appendix \ref{sec:appendix_results_mask_gains}.

\begin{table}[t]
\centering
\footnotesize
\begin{subtable}[t]{\columnwidth}
\centering
\resizebox{\columnwidth}{!}{%
\begin{tabular}{lcccccc}
\toprule
 & \multicolumn{3}{c}{COCO} & \multicolumn{3}{c}{Flickr30k} \\
Model & R@1 & R@5 & R@10 & R@1 & R@5 & R@10 \\
\midrule
CLIP & 35.33 & 59.97 & 70.15 & 65.20 & 87.24 & 92.00  \\
CLIP+masking & \textbf{37.47} & \textbf{62.06} & \textbf{72.25} & \textbf{66.96} & \textbf{88.56} & \textbf{93.18} \\
\bottomrule
\end{tabular}%
}
\caption{Image retrieval}
\label{tab:image_retrieval_sub}
\end{subtable}

\vspace{1em}

\begin{subtable}[t]{\columnwidth}
\centering
\resizebox{\columnwidth}{!}{%
\begin{tabular}{lcccccc}
\toprule
 & \multicolumn{3}{c}{COCO} & \multicolumn{3}{c}{Flickr30k} \\
Model & R@1 & R@5 & R@10 & R@1 & R@5 & R@10 \\
\midrule
CLIP & 56.06 & 79.48 & 86.84 & 85.10 & 97.30 & 99.00 \\
CLIP+masking & \textbf{57.74} & \textbf{79.96} & \textbf{87.40} & \textbf{87.40} & \textbf{97.90} & \textbf{99.10} \\
\bottomrule
\end{tabular}%
}
\caption{Text retrieval}
\label{tab:text_retrieval_sub}
\end{subtable}
\vspace{-1em}
\caption{Zero-shot retrieval results for pretrained CLIP ViT L-14 on COCO \cite{chen2015coco} and Flickr30k \cite{young2014flickr} show performance gains from masking sink tokens at the final layers.}
\label{tab:combined_retrieval}
\end{table}


\section{Analysis of Sink Tokens}
\label{sec:mechanisms}

The efficiency and performance improvements demonstrated by our Fast Nystr\"om Attention (Section~\ref{sec:fna}) method and masking-based denoising method (Section~\ref{sec:performance_efficiency}) stem from fundamental mechanisms governing token interactions in vision transformers. In particular, we identify \emph{mutual suppression} among tokens as the driving force behind the emergence of massive and artifact tokens. Below, we detail how this suppression shapes attention dynamics, underpins efficient approximations, and motivates our masking strategy.

\subsection{Mechanisms of Token Suppression}

Our experiments show that massive tokens emerge through a distinct \emph{phased} progression, driven by mutual suppression and magnified by MLP layers. Although the following layer indices refer specifically to CLIP ViT L-14, the same qualitative patterns arise in other pretrained ViTs.

\paragraph{Emergence Phase (Layers 9--10).}
In the early-to-mid layers (e.g., layers 9 and 10 in CLIP ViT L-14), tokens with slightly larger activations begin to exhibit suppressive behaviors toward other potential sink tokens (including themselves). These suppressive signals, when passed through the MLP’s nonlinear transformations, create a compounding feedback loop. Once a token has a slight size advantage, it increasingly dampens its competitors’ growth while growing itself. Concretely, in a multi-head self-attention block, each token $j$ (serving as a \emph{key}) contributes a rank-$H$ subspace
\begin{align}
    \Sl{\ell}_j = \text{span}(\cbn{\Vl{\ell}_{h, j}}_{h \in [H]}) \subspace \R^d
\end{align}
which influences other tokens (as \emph{queries}) via the weighted combination from attention. When a (potentially) massive token $j$ has a large norm, its subspace $\mathcal{S}_j^{(\ell)}$ tends to \emph{negatively} project onto other tokens in attention, yielding a \emph{destructive} or suppressive effect on their subsequent activations.

\paragraph{Consolidation Phase (Layers 11--12).} 
The largest tokens fully mature into massive tokens in layers 11 and 12, absorbing a disproportionately large share of attention from all other tokens. Since attention weights are nonnegative (post-softmax), the dominant tokens effectively channel attention values in a way that \emph{further} suppresses remaining mid-sized contenders. Mathematically, if $P_{i,j}$ measures the mean normalized projection of token $j$’s subspace onto $i$:
\begin{align}
    P_{i, j}
    = \frac{1}{H} \dss{h}{} \frac{\innern{\emb{\ell}{i}, \Vl{\ell}_{h, j}}}{\normn{\emb{\ell}{i}}}
    = \inner{\frac{\emb{\ell}{i}}{\normn{\emb{\ell}{i}}}, \frac{1}{H} \dss{h}{} \Vl{\ell}_{h, j}},
\end{align}
then a \emph{strongly negative} $P_{i,j}$ indicates $j$ significantly \emph{suppresses} $i$. In layers 9--10, the potential sink tokens heavily penalize each other’s growth; by layers 11--12, a small number of them have ``won'' the competition and become the new attention sinks as seen in Appendix \ref{sec:appendix_analysis} Figure \ref{fig:massive_token_evolution}.

\begin{figure*}[!t]
    \centering
    \includegraphics[width=1\linewidth]{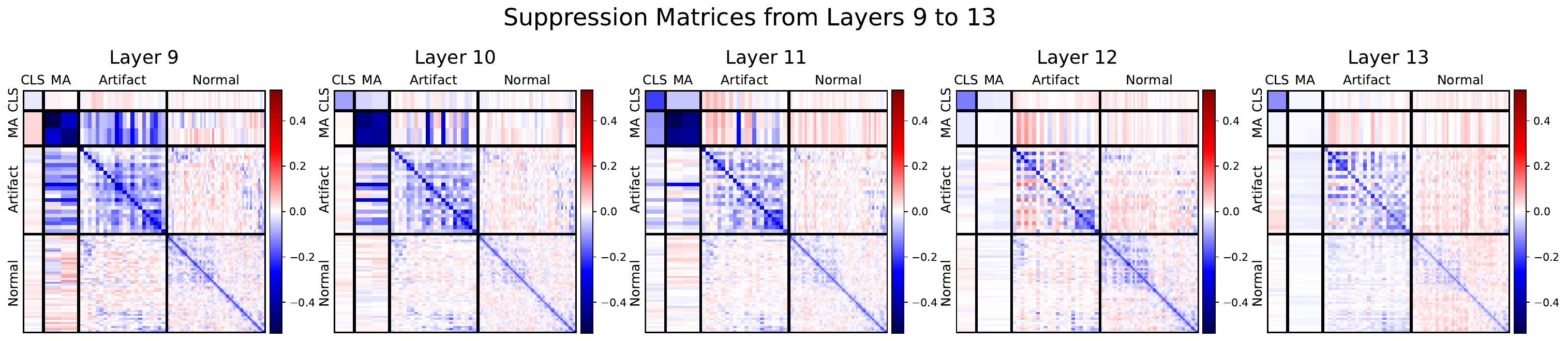}
    \vspace{-1.5em}
    \caption{We observe in layers 9 and 10 that the pairwise suppression within the set of (potential) sink tokens is particularly strong. This effect decays and is absent at layers 12 and beyond. We also observe that the attention projection of the subspaces of massive tokens onto other tokens is particularly small after layer 12, which is consistent with findings that the values of massive tokens post-emergence are significantly smaller than average. It is also interesting to note that each token's projection onto itself is strongly negative, suggesting that its attention to itself may be partially destructive.}
    \label{fig:suppresion_transition}
\end{figure*}

\paragraph{Stabilization Phase (Layer 13+).}
Past layer 12, the suppression mechanism ceases while the massive tokens remain stably large, acting as bottlenecks for global information flow but no longer contending with other latent sink tokens.

\subsection{Implications for Efficient Attention}


The structured hierarchy of token importance revealed by our analysis—where (1) CLS tokens provide global context, (2) massive tokens dominate local attention patterns, and (3) artifact tokens represent latent redundancy—directly informs the design of Fast Nyström Attention. By recognizing these key roles, we can strategically sample tokens for Nyström approximation without compromising attention fidelity. The mutual suppression dynamics ensure that FPS naturally selects these critical tokens, as they occupy distinct regions of the feature manifold (Figure \ref{fig:dino_clip_artifacts}).

\subsection{Theoretical Underpinnings of Masking Gains}
Masking sink tokens in the later layers consistently boosts performance by rebalancing the attention dynamics toward more informative normal tokens. In tasks like classification or retrieval, the CLS token relies on attending to normal tokens for a rich global image representation. Sink tokens siphon attention from these normal tokens, degrading the CLS token’s ability to aggregate global information in the final layers. By masking sink tokens, we free the CLS token to attend more effectively to the meaningful patches, improving its final-layer representation. For segmentation or other dense tasks, we apply a linear probe to the final-layer patch embeddings. Sink tokens disrupt local coherence by overpowering normal tokens in attention. Masking preserves spatial fidelity yielding better dense predictions. While removing established massive tokens can trigger artifact tokens to grow, this process occurs earlier in the network. In the final layers, masking eliminates interfering activations without reintroducing new ones, denoising the final representation.

\section{Conclusion}
\label{sec:conclusion}
Our work reveals that the emergent phenomena of massive and artifact tokens in vision transformers govern the information flow through attention mechanisms and present an opportunity for efficiency gains. By introducing Fast Nystr\"om Attention (FNA), a training-free approach that exploits these token properties for linear-time, low-rank approximations of self-attention, we demonstrate significant reductions in computational and memory overhead while preserving competitive performance on a variety of downstream tasks. Our comprehensive analysis—spanning iterative and non-iterative detection methods as well as the strategic masking of attention sinks—sheds light on the underlying suppression dynamics that shape token interactions in attention, enabling us to enhance global feature aggregation and improve downstream tasks.


{\small
\bibliographystyle{ieee_fullname}
\bibliography{11_references}
}

\clearpage
\appendix
\section{Additional Results}
\label{sec:appendix_results}

\subsection{FNA Sampling Configurations}
\label{sec:appendix_results_fna_sampling}

\begin{figure}[b]
    \centering
    \begin{subfigure}[t]{1.0\linewidth}
        \centering
        \includegraphics[width=\linewidth]{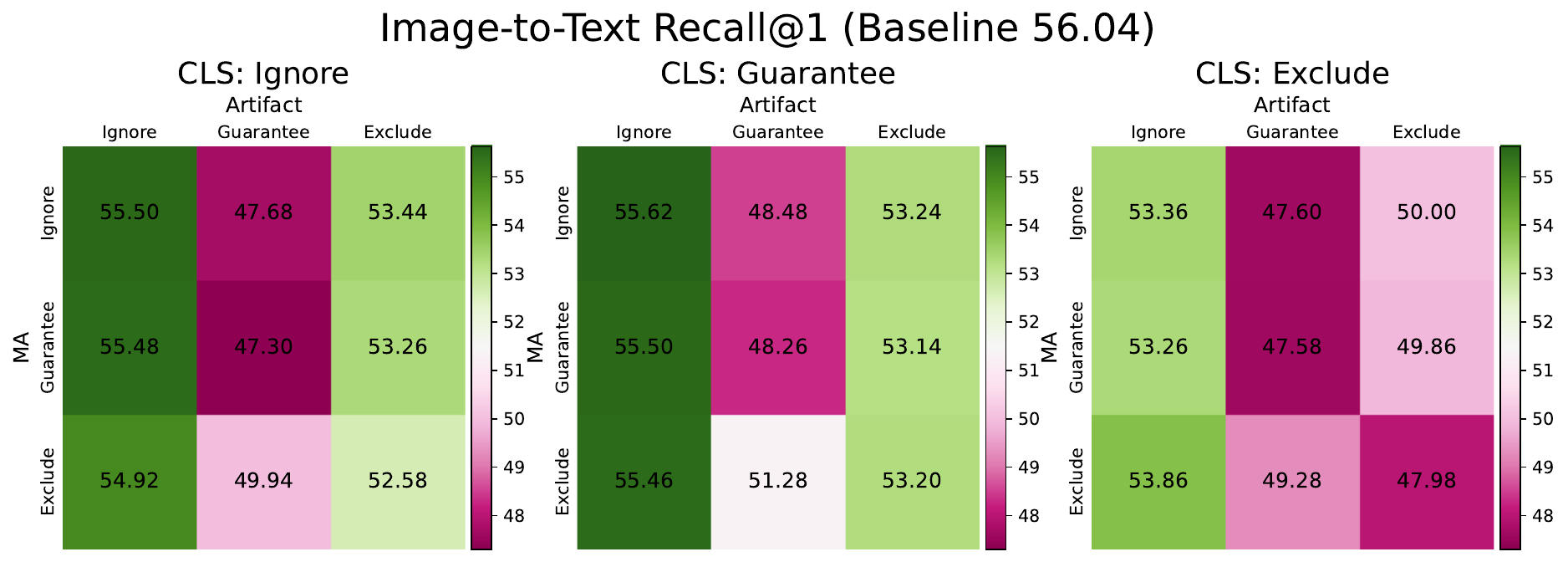}
        \label{fig:image_to_text_grid_a}
    \end{subfigure}
    \hfill
    \begin{subfigure}[t]{1.0\linewidth}
        \centering
        \includegraphics[width=\linewidth]{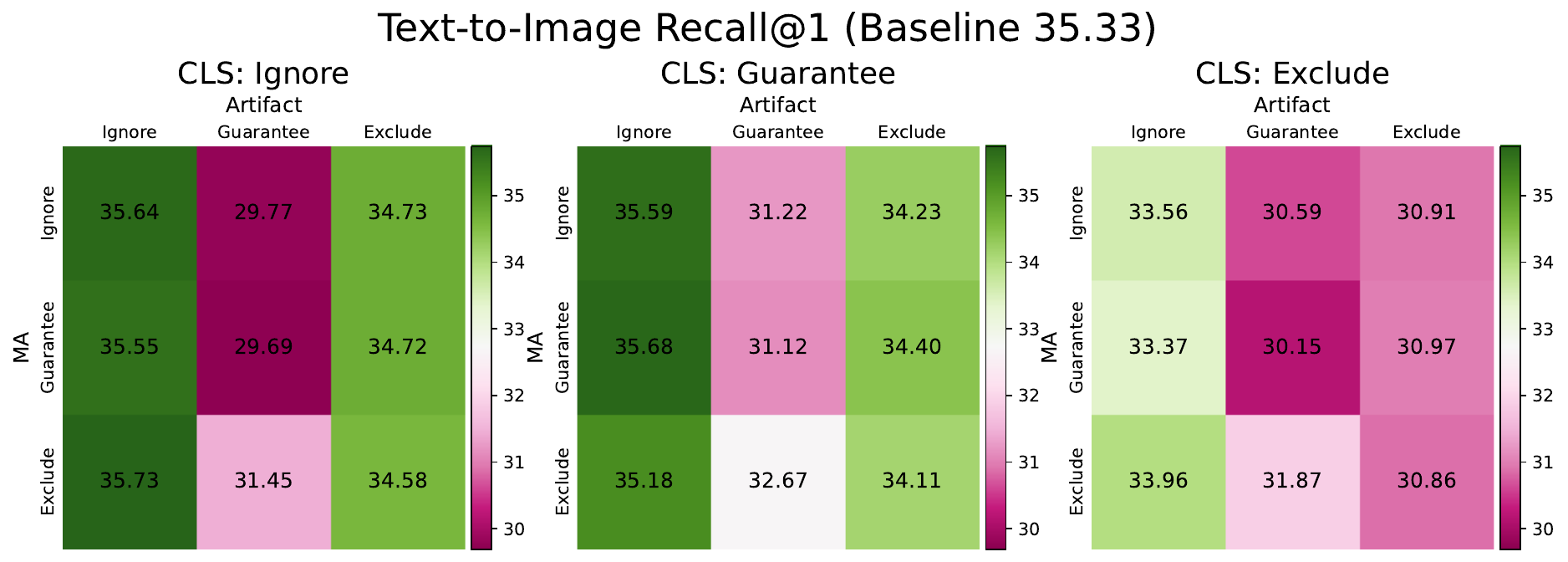}
        \label{fig:image_to_text_grid_b}
    \end{subfigure}
    \vspace{-1em}
    \caption{COCO retrieval metrics on all $3^3 = 27$ FPS sampling configurations for image-to-text (top) and text-to-image (bottom).}
    \label{fig:coco_recall_subfigs}
\end{figure}

Using pretrained CLIP ViT-L14 \cite{radford2021clip}, we perform a grid search on all $3^3 = 27$ different combinations of ignoring, guaranteeing, and excluding CLS, massive, and artifact tokens when sampling for Fast Nystr\"om Attention (Figure \ref{fig:coco_recall_subfigs}). We find that solely guaranteeing the CLS token performs nearly identically to guaranteeing the sampling of massive tokens, significantly better than guaranteeing the sampling of artifact tokens, and notably better than excluding either. 

\begin{figure}[b]
    \centering
    \includegraphics[width=1\linewidth]{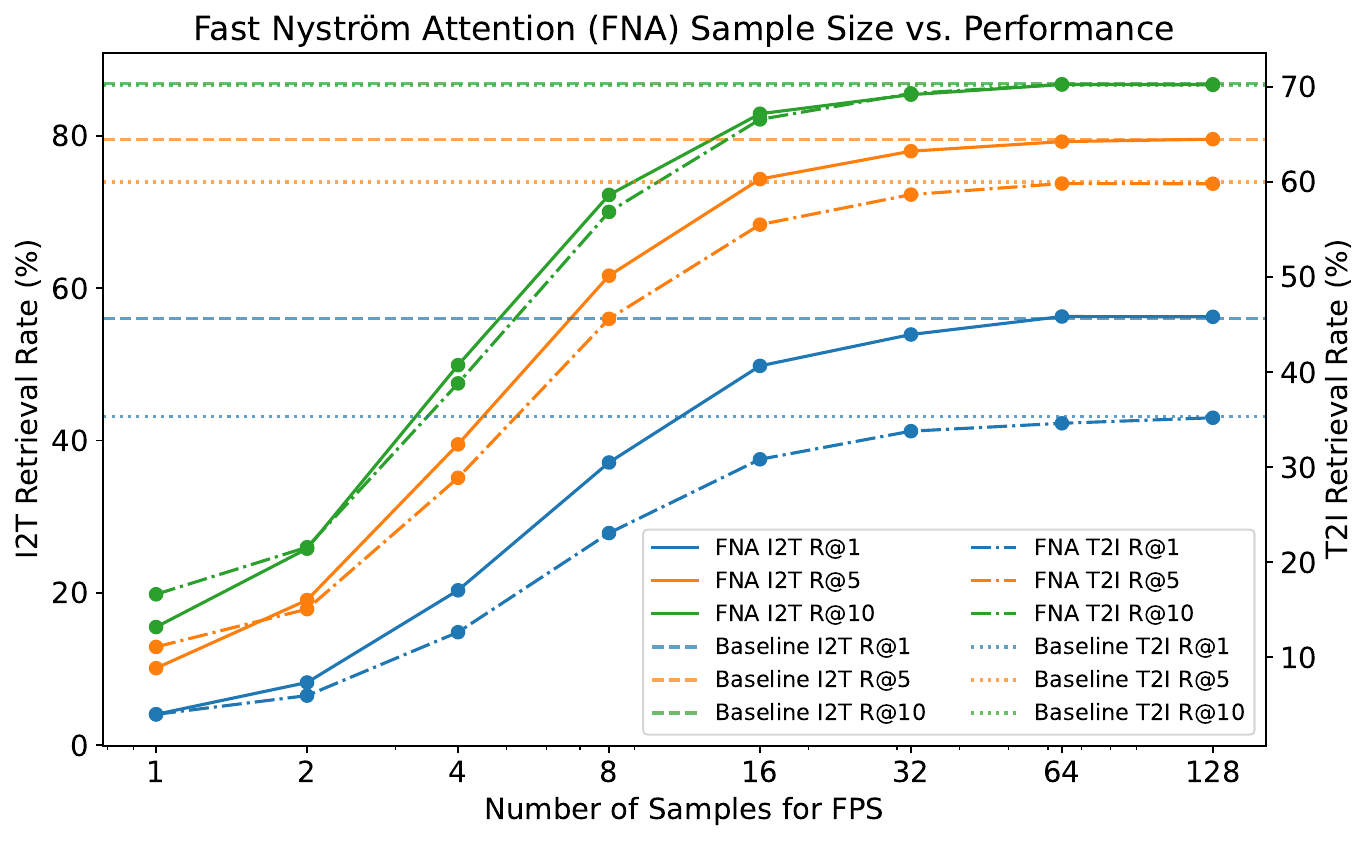}
    \vspace{-1em}
    \caption{Furthest point sampling (FPS) sample size vs. performance for COCO retrieval with CLIP ViT L-14 at 224x224px input resolution.}
    \label{fig:fna_sample_size}
\end{figure}

Figure \ref{fig:fna_sample_size} shows evaluation on COCO \cite{chen2015coco} retrieval with different sample sizes used for Nystr\"om approximation. Sampling > 32 tokens gives nearly identical performance to standard attention.

\subsection{Finetuning Comparison with Existing Linear Attention Methods}
\label{sec:appendix_results_linear}

We compare training efficiency of Fast Nystr\"om Attention with existing linear attention methods \cite{choromanski2020performer} \cite{wang2020linformer} \cite{xiong2021nystromformer} by finetuning CLIP ViT L-14 for one epoch on COCO and evaluating retrieval. Validation metrics reported in Table \ref{tab:linear_attn} demonstrate improved performance in both training free and finetuning settings.

\begin{table}[t]
    \centering
    \footnotesize
    \resizebox{0.9\columnwidth}{!}{%
        \begin{tabular}{lccccc}
            \toprule
             & & \multicolumn{2}{c}{Text-to-Image} & \multicolumn{2}{c}{Image-to-Text} \\
            Model & Finetuned & R@1 & R@5 & R@1 & R@5 \\
            \midrule
            Baseline & \xmark & 35.33 & 59.97 & 56.06 & 79.42 \\
            Linformer \cite{wang2020linformer} & \xmark & 0.06 & 0.26 & 0.02 & 0.22 \\
            Performer \cite{choromanski2020performer} & \xmark & 4.28 & 12.07 & 3.76 & 11.06 \\
            FNA+seg. means \cite{xiong2021nystromformer} & \xmark & 32.96 & 57.80 & 49.32 & 74.18 \\ 
            \textbf{FNA+FPS (ours)} & \xmark & 35.91 & 60.43 & 56.52 & 79.32 \\
            \midrule
            Baseline & \cmark & 49.57 & 74.91 & 65.42 & 86.80 \\
            Linformer & \cmark & 18.51 & 42.40 & 22.56 & 47.48 \\
            Performer & \cmark & 41.80 & 69.07 & 53.48 & 79.32 \\
            FNA+seg. means & \cmark & 45.45 & 71.75 & 60.92 & 83.66 \\ 
            \textbf{FNA+FPS (ours)} & \cmark & 48.40 & 73.98 & 64.48 & 86.24 \\ 
            \bottomrule
        \end{tabular}%
    }
    \caption{Validation retrieval performance on COCO using pretrained CLIP ViT L-14 with different linear attention methods applied. We finetune only the QKV projection layers (and down-projection in Linformer). For consistency, we set the down-projection dimension in Linformer and the sample size in Performer and FNA  to 64. We finetune each model with identical hyperparameters on 1 epoch of the COCO training set.}
    \label{tab:linear_attn}
\end{table}

\subsection{Qualitative Results for LLaVa Inference}
\label{sec:appendix_results_llava}

\begin{figure*}
    \centering
    \includegraphics[width=1\linewidth]{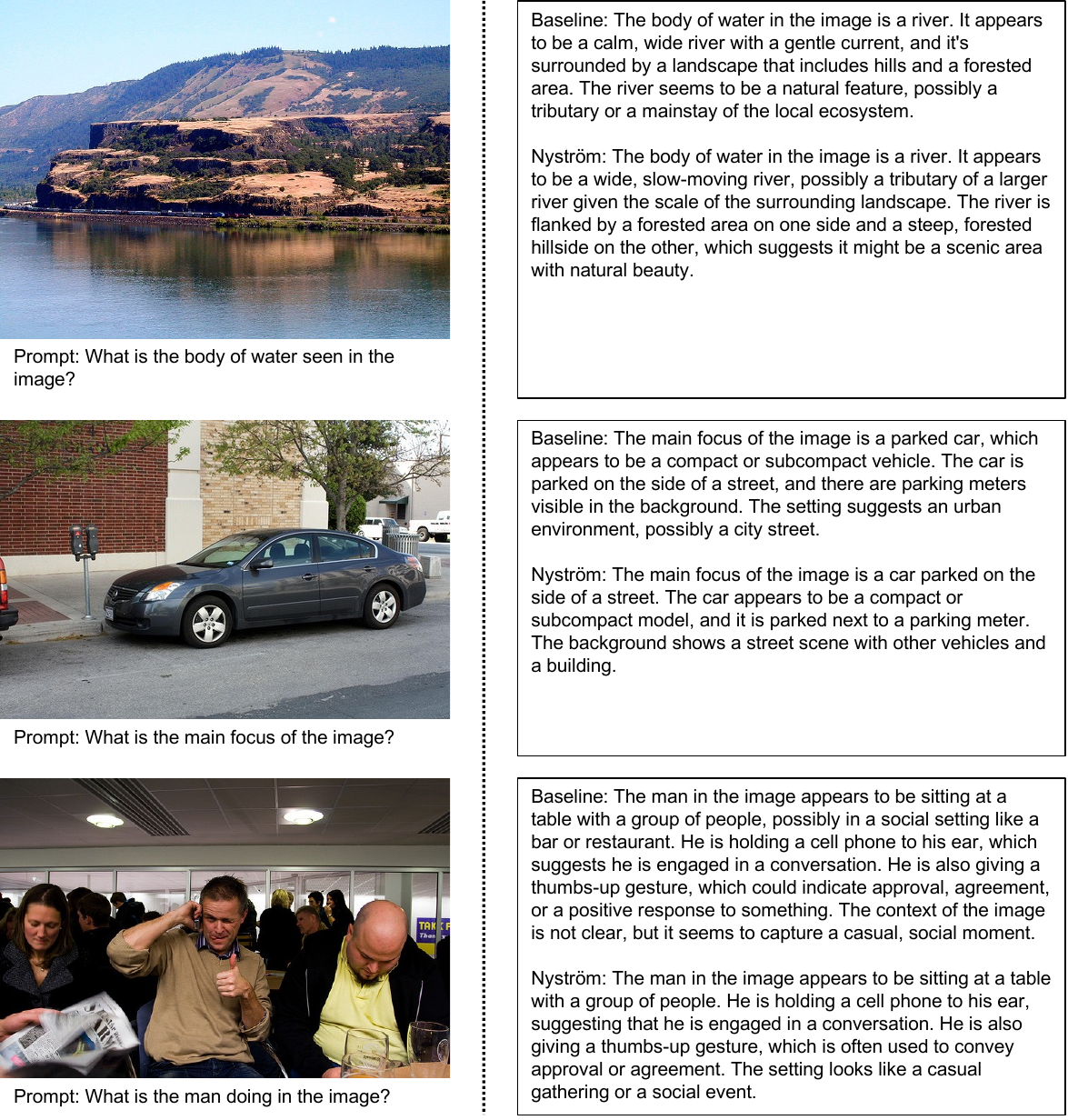}
    \caption{Example outputs generated by LLaVA-NeXT-7B \cite{liu2023llava} using input images and text prompts from the COCO VQA \cite{AgrawalCOCOVQA} dataset. We apply Fast Nyström Attention (layers 18 to 32, sample size = 64) to the image tokens in the LLaMA backbone. Greedy decoding is used for generation when comparing against the baseline.}
    \vspace{128in}
    \label{fig:llava_qualitative}
\end{figure*}

Figure \ref{fig:llava_qualitative} shows qualitative examples of responses generated by LLaVA-NEXT-7B \cite{liu2023llava} on COCO VQA \cite{AgrawalCOCOVQA} prompts, using our Fast Nyström Attention method. These examples illustrate that our approach preserves the semantic quality of the answers while reducing the computational cost.

\subsection{Sink Token Masking Ablation}
\label{sec:appendix_results_mask_ablation}

\begin{figure}
    \centering
    \begin{subfigure}[t]{1.0\linewidth}
        \centering
        \includegraphics[width=\linewidth]{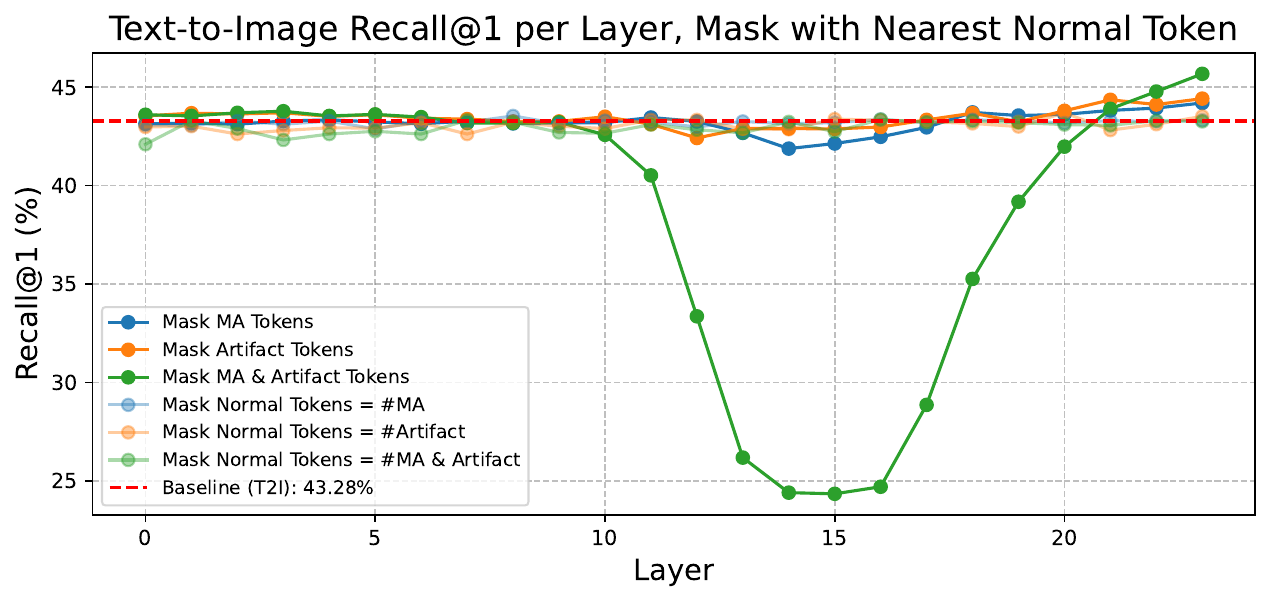}
        \label{fig:t2i_replace}
    \end{subfigure}
    \vspace{0.1cm} 
    \begin{subfigure}[t]{1.0\linewidth}
        \centering
        \includegraphics[width=\linewidth]{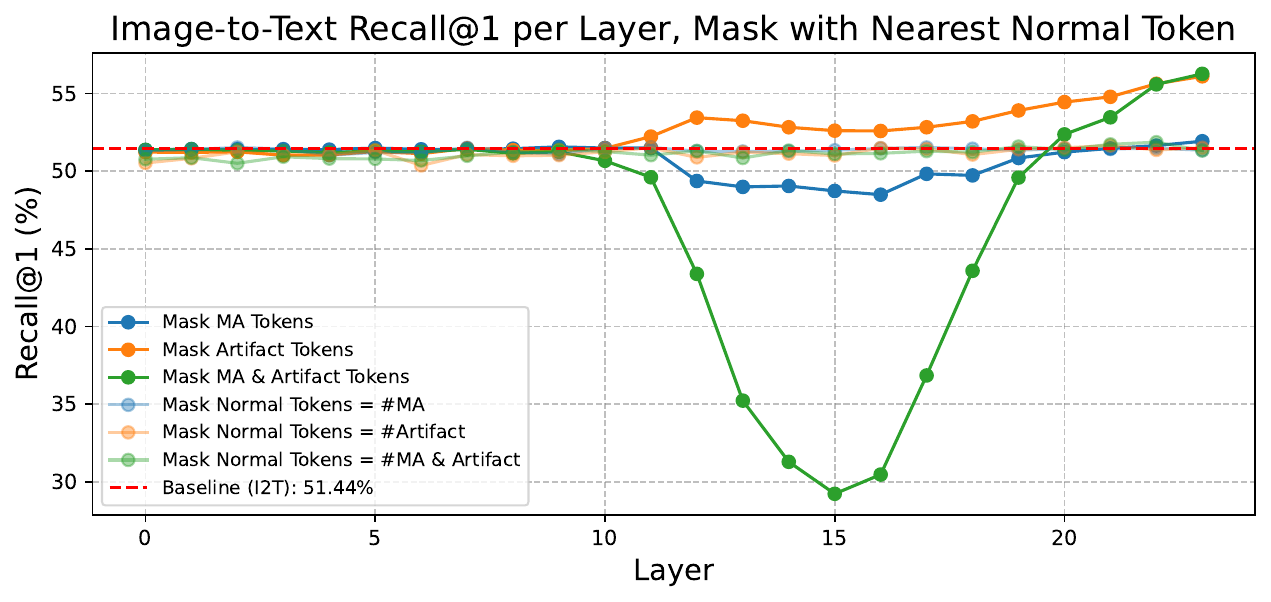}
        \label{fig:i2t_replace}
    \end{subfigure}
    \vspace{-1em}
    \caption{Zero-shot image and text retrieval ablation performed on COCO with pretrained CLIP ViT L-14 show the effect of masking sink tokens at each layer.}
    \label{fig:retrieval_ablation}
\end{figure}

In Figure \ref{fig:retrieval_ablation}, we analyze the role sink tokens play at each layer in CLIP ViT-L14 by selectively masking them with the nearest normal token neighbor. When masking sink tokens tokens prior to their formation (\textit{i.e.} masking proto-sink tokens), performance is unaffected as another subset of normal tokens becomes sink tokens. Masking MA tokens after their formation drops performance incrementally since artifact tokens can become massive if needed. Notably, only after removing both MA and artifact tokens at this stage does performance drops considerably, supporting both the importance of sink tokens and the redundant nature of artifact tokens. Removing sink tokens at later layers boosts retrieval metrics as presented in Section \ref{sec:performance_efficiency} and Appendix \ref{sec:appendix_results_mask_gains}.

\subsection{Masking Gains on Classification and Segmentation}
\label{sec:appendix_results_mask_gains}

Section \ref{sec:performance_efficiency} demonstrates how masking out massive and artifact tokens in the final layer of pretrained CLIP improves performance on retrieval tasks. Similarly, this masking strategy yields a small boost in zero-shot ImageNet accuracy (Table~\ref{tab:vision_only_results}) with both CLIP and DINOv2. For dense prediction tasks such as semantic segmentation on VOC2012~\cite{pascal-voc-2012} and ADE20k~\cite{zhou2017ade}, masking massive and artifact tokens likewise produces cleaner, more coherent patch features (Figure~\ref{fig:dino_clip_artifacts}) and translates to a minor improvement in segmentation performance.

\begin{table}
\centering
\footnotesize
\resizebox{\columnwidth}{!}{%
\begin{tabular}{lcccccc}
\toprule
 & \multicolumn{2}{c}{ImageNet} & \multicolumn{2}{c}{VOC2012} & \multicolumn{2}{c}{ADE20k} \\
Model & Top1 & Top5 & aACC & mIoU & aACC & mIoU \\
\midrule
CLIP & 75.96 & 94.82 & 90.33 & 66.60 & 69.55 & 34.84 \\
CLIP+masking & \textbf{76.26} & \textbf{94.86} & \textbf{90.59} & \textbf{66.96} & \textbf{69.91} & \textbf{35.09} \\
\midrule
DINOv2 & \textbf{78.62} & 92.91 & 94.12 & 77.54 & 78.65 & 44.48 \\
DINOv2+masking & \textbf{78.62} & \textbf{93.06} & \textbf{94.99} & \textbf{79.65} & \textbf{78.76} & \textbf{44.72} \\
\bottomrule
\end{tabular}%
}
\caption{Vision-only results for pretrained CLIP and DINOv2 ViT L-14. ImageNet \cite{deng2009imagenet} classification evaluation is performed in the zero-shot setting. Segmentation on VOC2012 \cite{pascal-voc-2012} and ADE20k \cite{zhou2017ade} is performed via fitting linear probes to output of the final layers. We show minor but consistent performance gains from masking sink tokens in the final layers.}
\label{tab:vision_only_results}
\end{table}
\section{Analysis}
\label{sec:appendix_analysis}

\begin{figure}[H]
    \centering
    \includegraphics[width=1\linewidth]{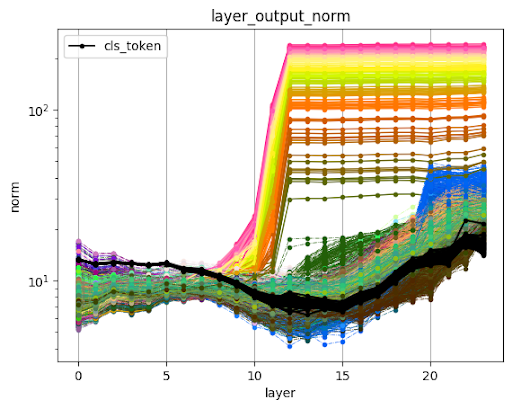}
    \vspace{-1em}
    \caption{Plot of activation norms of tokens across 50 images over all layers of CLIP ViT-L14 show that the massive tokens become large primarily in layers 11 and 12.}
    \label{fig:massive_token_evolution}
\end{figure}

\subsection{Definitions}
\label{sec:definitions}
Similar to Section \ref{sec:performance_efficiency}, the layer indices used in this section refer specifically to CLIP ViT L-14; however, the same qualitative patterns arise in other large pretrained ViTs. We define two key operations that we will use to analyze the formation of massive tokens:
\begin{definition}[Masking]
    For a vector $w \in \R^n$ and mask $m \in \cb{0, 1}^n$, we define $\sfm$ as \begin{align}
        \sfm(w, m) = \sf(w - \infty \cdot (1 - m)).
    \end{align} In other words, \begin{align}
        \sfm(w, m)_i = \begin{cases}
            \frac{e^{w_i}}{\dss{m_j = 1}{} e^{w_j}}   & \text{if} \ m_i = 1  \\
            0   & \text{otherwise}
        \end{cases}.
    \end{align}
\end{definition}

\begin{definition}[Sinking]
    For a matrix $w \in \R^n$ and mask $m \in \cb{0, 1}^n$, we define $\sfs$ as \begin{align}
        \sfm(w, m) = m \odot \sf(w).
    \end{align} In other words, \begin{align}
        \sfs(w, m)_i = \begin{cases}
            \frac{e^{w_i}}{\dss{j}{} e^{w_j}}   & \text{if} \ m_i = 1  \\
            0   & \text{otherwise}
        \end{cases}.
    \end{align}
\end{definition}

For operand tensors of multiple dimensions, these operations similarly to $\sf$ will only be relevant on the last dimension, while any precedent dimensions will be interpreted as batch dimensions. As a result, we introduce the notations $\attnm{\ell}(\cdot, M)$ and $\attns{\ell}(\cdot, M)$ for the application of Equation \ref{eqn:attn} with the substitution of $\sf(\cdot)$ for $\sfm(\cdot, M)$ and $ \sfs(\cdot, M)$ respectively, as well as $\lyrm{\ell}(\cdot, M)$ and $\lyrs{\ell}(\cdot, M)$ to similarly substitute Equation \ref{eqn:layer}. \\

While masking is common operation frequently used to enable causal masking, pad of heterogeneous sequences, and regularize training, the effect of masking on when applied to a pretrained model is not intuitively clear due to the global rescaling of the attention vector. I.e. for value vector $v \in \R^n$, \begin{align}
    \tr{\sfm(w, m)}v
    & = \frac{\dss{j}{} e^{w_j}}{\dss{m_j = 1}{} e^{w_j}} \tr{(m \odot \sf(w))}v   \\
    & = \frac{\dss{j}{} e^{w_j}}{\dss{m_j = 1}{} e^{w_j}} \dss{m_i = 1}{} \sf(w)_iv_i    \\
    & = \frac{\dss{j}{} e^{w_j}}{\dss{m_j = 1}{} e^{w_j}} (\tr{\sf(w)}v - \dss{m_i = 0}{} \sf(w)_iv_i).
\end{align} Due to the change in the exponential sum, the rescaling of the attention vector may not impact the computational path in a small way, especially if attention weight to a masked token is large which we will observe later. Thus, we identify sinking as a useful intermediate that allows us to study in specific the effects of the additive signal transmitted to a token as a result of the attention mechanism, as we have \begin{align}
    \tr{\sfs(w, m)}v
    & = \tr{\sf(w)}v - \dss{m_i = 0}{} \sf(w)_iv_i.
\end{align}

\begin{figure}[t]
    \centering
    \includegraphics[width=1\linewidth]{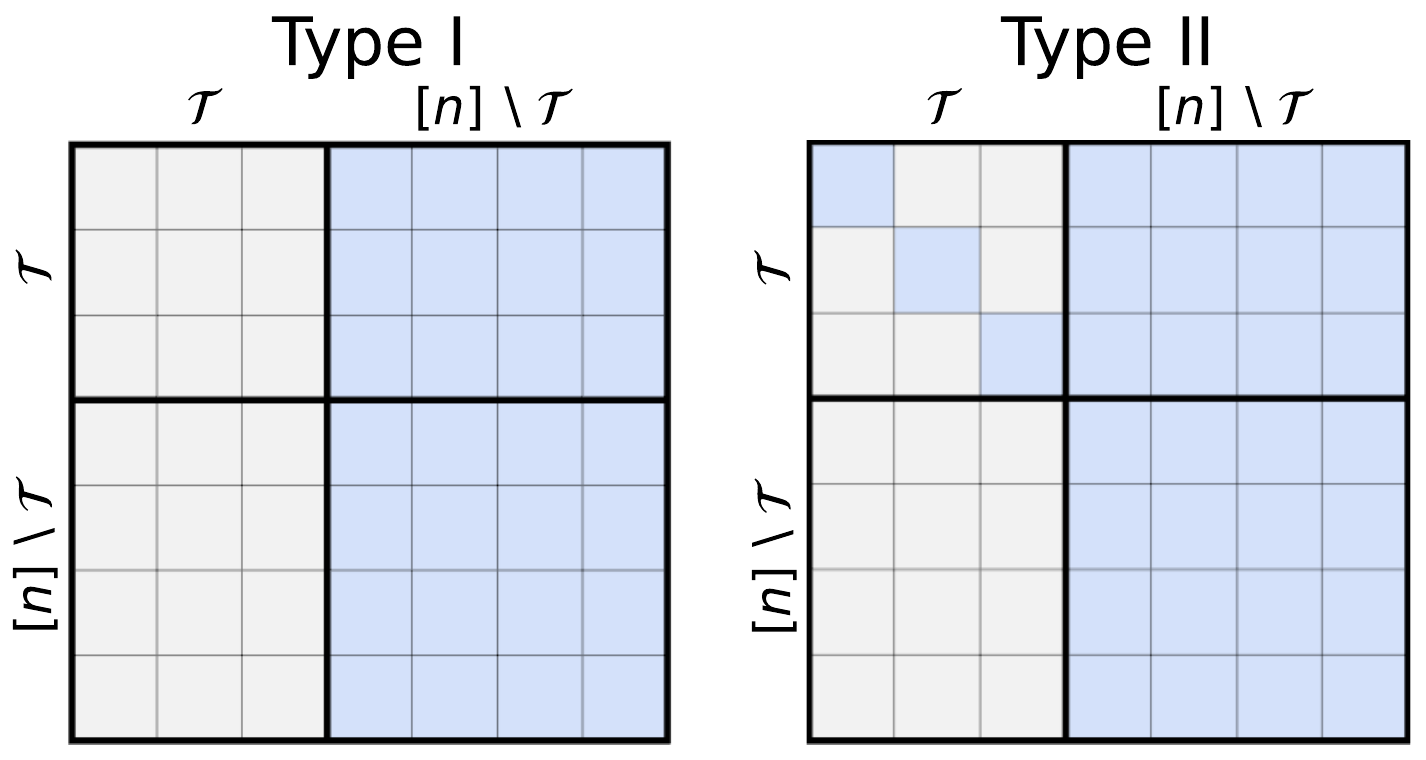}
    \caption{Different masking patterns with respect to the interest set of tokens $\mc{T}$ where blue represents token query-key pairs that are allowed while gray represents query-key pairs that are disallowed.}
    \label{fig:masking_patterns}
\end{figure}

We identify two masking patterns that will be useful in our analysis of massive token activations, where the patterns regard a set of interest tokens $\mc{T} \subseteq [n]$ with $\mc{T}$ being generally small. \begin{itemize}
    \item We define the \textbf{Type I} masking pattern $M_I(\mc{T})$ as $M$ where $m_{i, j} = \indic{j \notin \mc{T}}$. This means that no token will attend to any token in $\mc{T}$.


    \item We define the \textbf{Type II} masking pattern $M_{II}(\mc{T})$ as $M$ where $m_{i, j} = \indic{i = j \lor j \notin \mc{T}}$. This means that for all tokens $t \in \mc{T}$, only $t$ can attend to $t$.
\end{itemize}
These masking patterns are depicted in Figure \ref{fig:masking_patterns}. We refer to the replacement of the $\sf(W)$ operation with $\sfs(W, M_I(\mc{T}))$ as ``\textbf{Type I} sinking set $\mc{T}$'' with identical colloquialism enjoyed for \textbf{Type II}. Furthermore, the particular interest sets that we will be applying the masking patterns to will be made explicit in Section \ref{sec:mechanisms}.

\begin{figure*}[t]
    \centering
    \begin{subfigure}[t]{0.49\linewidth}
        \centering
        \includegraphics[width=\linewidth]{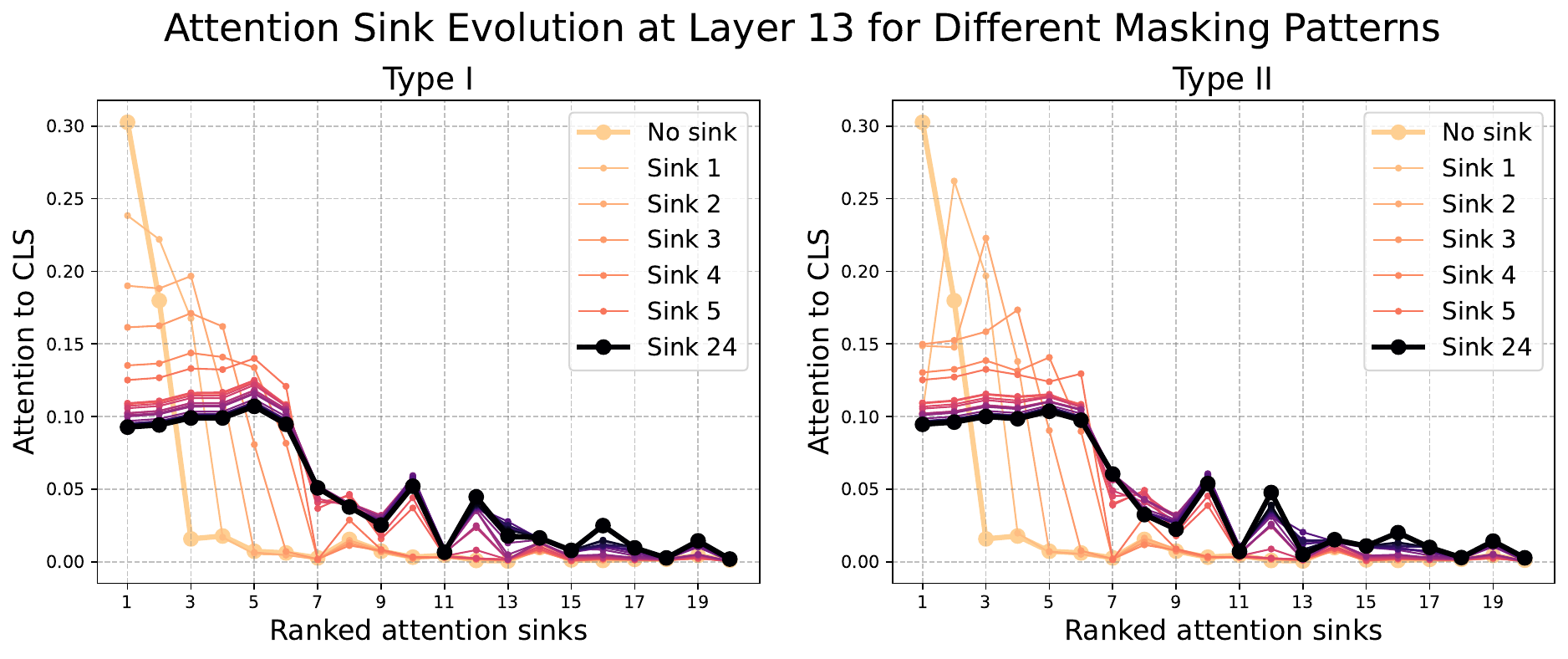}
        \caption{}
        \label{fig:attention_sink_decay}
    \end{subfigure} %
    \begin{subfigure}[t]{0.49\linewidth}
        \centering
        \includegraphics[width=\linewidth]{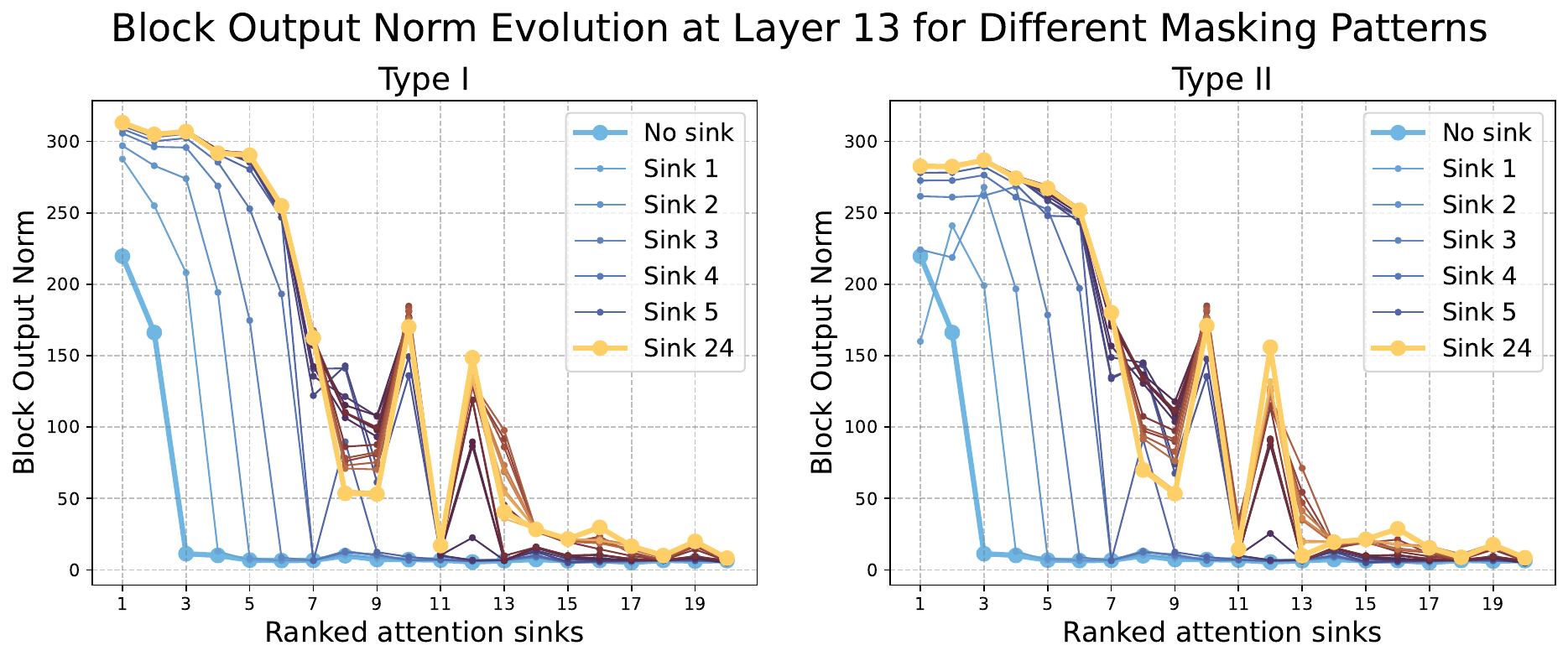}
        \caption{}
        \label{fig:layer_output_decay}
    \end{subfigure}
    \caption{The $x$-axis shows individual tokens that are ranked by order of removal in the iterative masking procedure, with the same sequence of top 20 tokens showed for all subplots. The $y$-axis of subfigure \ref{fig:attention_sink_decay} denotes the (unsunk) attention from the CLS token which we use as a more distinct proxy for incoming attention, while the $y$-axis of subfigure \ref{fig:layer_output_decay} denotes the block output magnitude which we determine to be strongly representative of the attention logits. While masking formally redistributes the attention pool across the remaining tokens, sinking can alternatively be interpreted as a zero-ing of values while maintaining the attention weights. Therefore, we observe that iterative masking results in the masked tokens attracting zero attention while subsequent sink tokens rise uncontested. On the other hand, iterative sinking allows sunk tokens the opportunity to ``retain their place'' in the attention distribution which we observe to be diminished but still significant.}
    \label{fig:combined}
\end{figure*}

\subsection{Facilitation of Largeness}
By running the transformer model without the attention mechanism in layers 9 to 12 (through either forwarding the block output $\Emb{\ell}$ direction to $\lyn{2}{\ell}$, or zeroing all attention values i.e. $\sfs(\Wl{\ell}, \mathbf{0})$), we identify the MLP in layers 11 and 12 as the main facilitators of largeness in massive tokens. However, we also observe that \begin{enumerate}[label=\arabic*)]
    \item removing the attention mechanism in layers 9, 10, 11, 12 result in some artifact tokens becoming massive that are not massive in the unmodified computational path, and

    \item for any interest set $\mc{T}$ (including $\emptyset$) that \textbf{Type I} masking or sinking in layers 9, 10, 11, 12 elicits the same set of massive tokens as \textbf{Type I} masking or sinking in layers 9, 10 and proceeding without attention in layers 11 and 12.
\end{enumerate} This suggests that while the MLP in layers 11 and 12 are the main driving force behind making tokens large, the attention mechanism in layers 9 and 10 determines which tokens become massive from a set of potential tokens that is determined by computation up to layer 8.

\begin{figure}[H]
    \centering
    \includegraphics[width=1\linewidth]{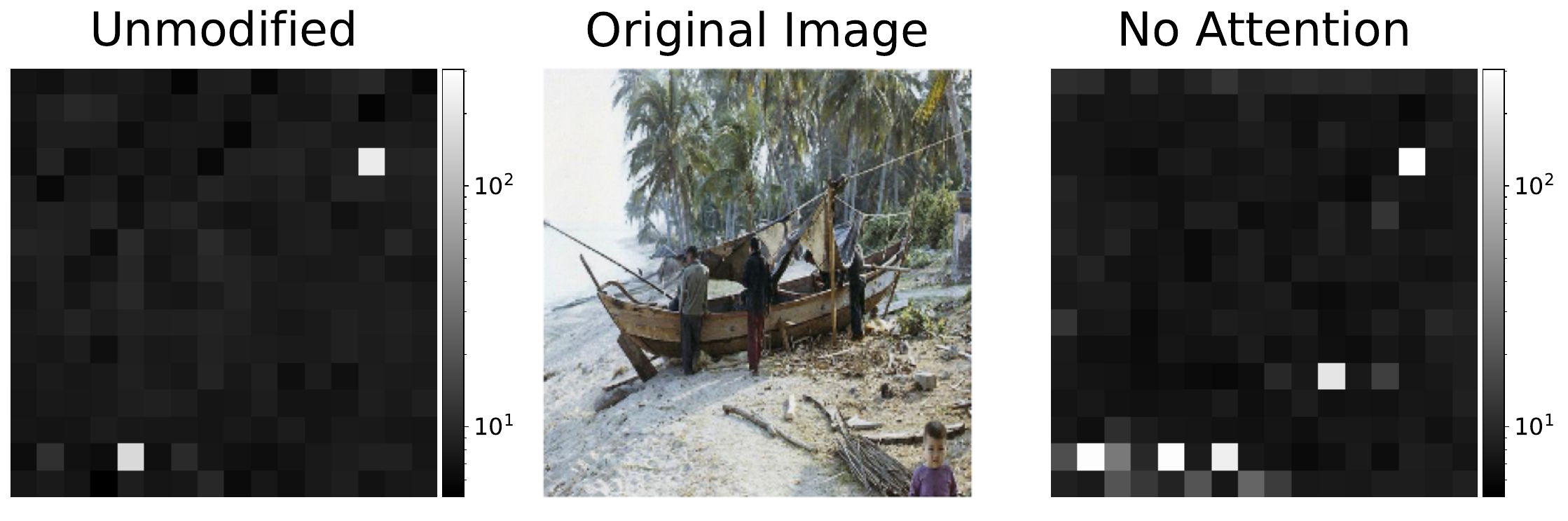}
    \caption{Running the model while ignoring the attention mechanism in layers 9, 10, 11, 12 result in some artifact tokens emerging as massive tokens that are not massive in the unmodified computational path.}
    \label{fig:mlp_growth}
\end{figure}

\subsection{Intra-Sink Signal Suppression}
\label{sec:suppression}

\subsubsection{Analysis of Type I Masking and Sinking}
We observed in Section \ref{sec:iterative_masking} that iterative masking of attention sinks results in substitute tokens becoming massive as well. However, the same can be said if we instead use iterative \textit{sinking}. However, the difference in results is that even when sinking, the sunk tokens retain their status as attention sinks, albeit diminished. We can see in Figure \ref{fig:attention_sink_decay} that iterative sinking of \textbf{Type I} results in gradual redistribution of attention while the sunk tokens still individually constitute notable fractions of incoming attention. However, we can also observe in Figure \ref{fig:layer_output_decay} that \textbf{Type I} sinking unilaterally increases the size of tokens for multiple iterations. This suggests that in a vacuum, each of the potential sink tokens emit a signal that negatively impacts the ability of other tokens to become large. Removing that signal via masking or sinking allows those tokens to grow. That the incoming attention to the newly sunk token decreases is a result of the saturation of lower-ranked tokens at large magnitudes that are ultimately bounded by Lipschitzness of the MLP.

\subsubsection{Comparison of Type I and Type II Sinking}
With the largest massive token denoted as $t_1$, we then consider what happens when we \textbf{Type II} sink $\cbn{t_1}$ at layers 9 and 10. Because the attention pattern of $t_1$ itself is untouched by \textbf{Type II} sinking $t_1$ alone, its value at the intermediate output of layer 9 is identical to that of unmodified computation. On the other hand, the attention pattern for any token $t \neq t_1$ is identical to that of \textbf{Type I} sinking. Because the MLP applies to individual tokens, we can say in short that the layer 9 output as a result of \textbf{Type II} sinking is unmodified for $t = t_1$, and equivalent to its counterpart in \textbf{Type I} for $t \neq t_1$ as well as larger than its counterpart in the unmodified path.

However, we observe that by the output of layer 10, token $t_1$ under \textbf{Type II} sinking is \textit{smaller} than its unmodified counterpart. Because its value has not changed as of the layer 9 output, this must result from attending to the secondary tokens that have become larger in sinking $t_1$ which immediately suggests that the largeness of the secondary tokens also comes with strengthened suppression signals. This reversal effect compounds up until layer 13 by which token $t_1$ is significantly smaller than its unmodified counterpart as seen in Figure \ref{fig:layer_output_decay}.


\clearpage
\onecolumn
\section{Model Zoo Visualization}
\label{sec:appendix_zoo}

\subsection{CLIP ViT L-14}

\begin{figure*}[!hbp]
    \centering
    \begin{subfigure}{1.0\linewidth}
        \centering
        \includegraphics[width=\linewidth]{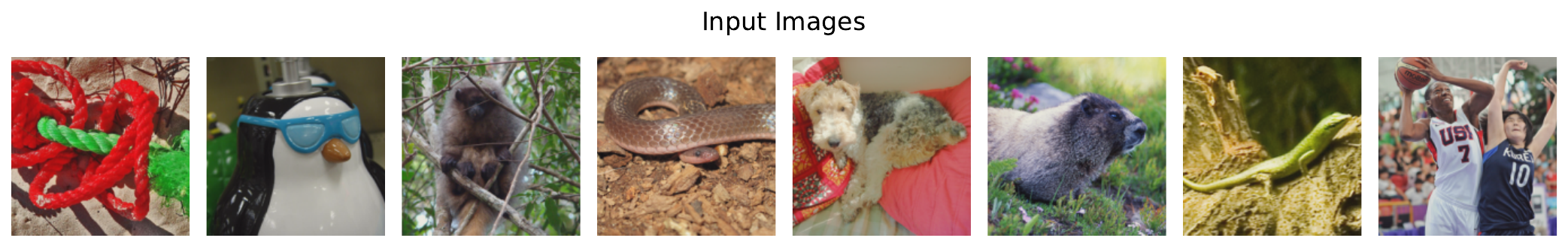}
        \label{fig:clip_0_input}
    \end{subfigure}

    \vspace{-8pt}
    \begin{subfigure}{1.0\linewidth}
        \centering
        \includegraphics[width=\linewidth]{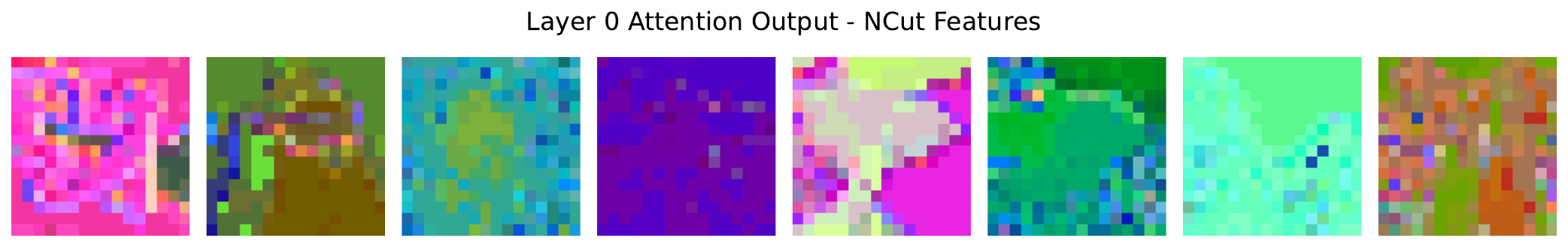}
        \label{fig:clip_0_attn}
    \end{subfigure}
    
    \vspace{-8pt}
    \begin{subfigure}{1.0\linewidth}
        \centering
        \includegraphics[width=\linewidth]{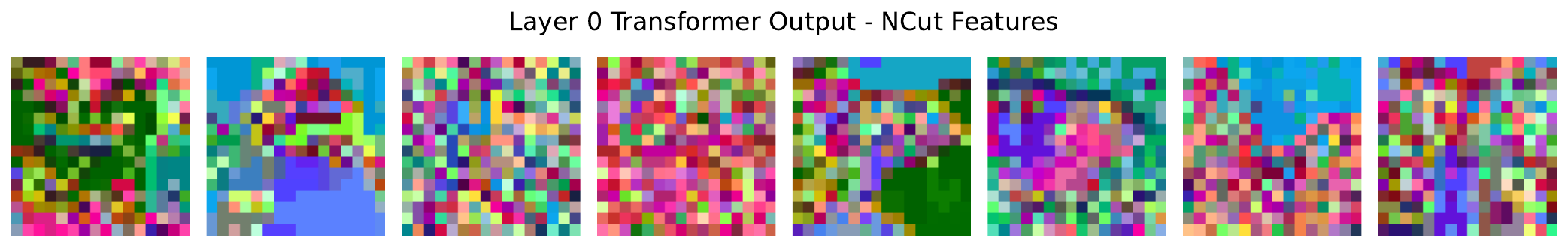}
        \label{fig:clip_0_block}
    \end{subfigure}
    
    \vspace{-8pt}
    \begin{subfigure}{1.0\linewidth}
        \centering
        \includegraphics[width=\linewidth]{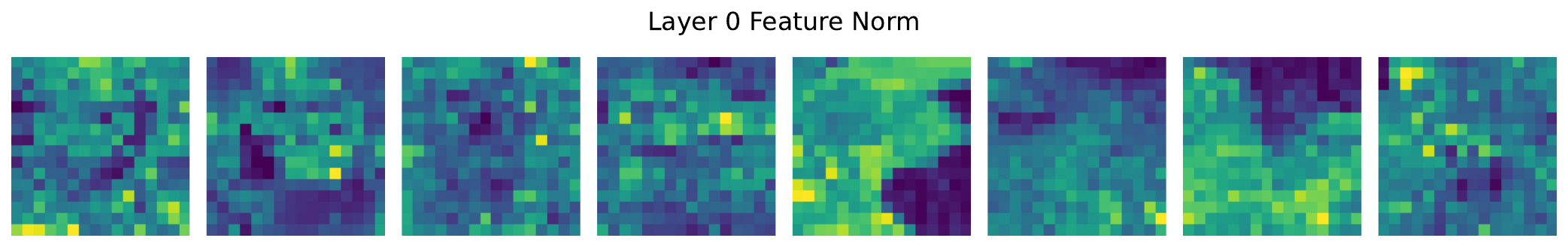}
        \label{fig:clip_0_norm}
    \end{subfigure}

    \vspace{-8pt}
    \begin{subfigure}{1.0\linewidth}
        \centering
        \includegraphics[width=\linewidth]{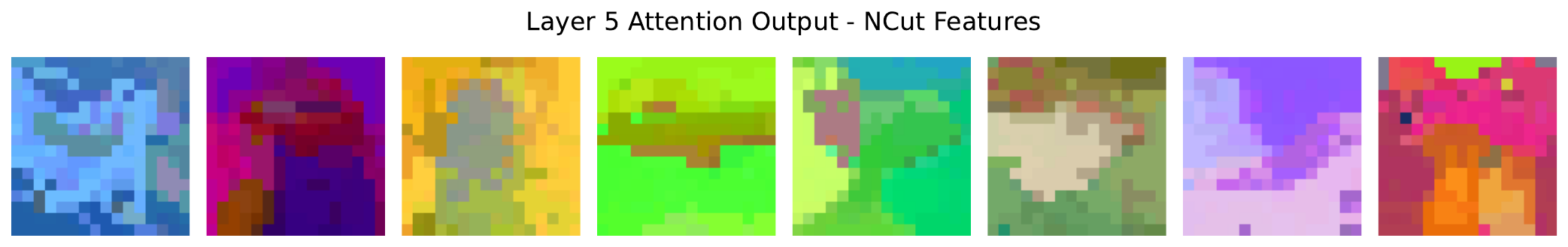}
        \label{fig:clip_5_attn}
    \end{subfigure}

    \vspace{-8pt}
    \begin{subfigure}{1.0\linewidth}
        \centering
        \includegraphics[width=\linewidth]{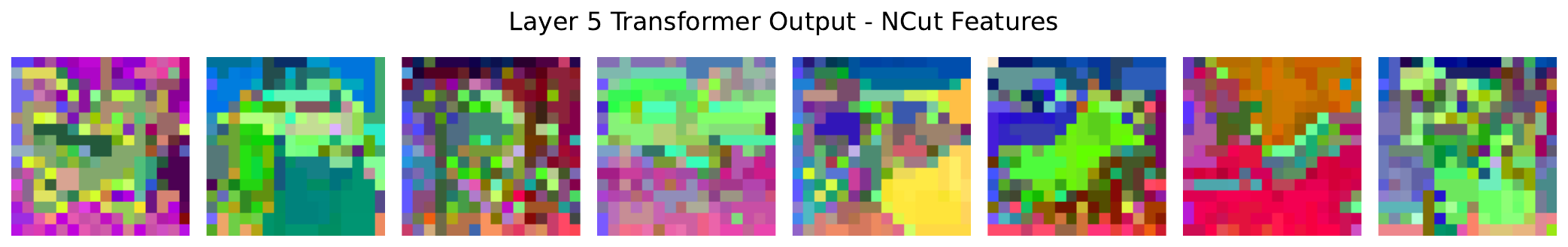}
        \label{fig:clip_5_block}
    \end{subfigure}

    \vspace{-8pt}
    \begin{subfigure}{1.0\linewidth}
        \centering
        \includegraphics[width=\linewidth]{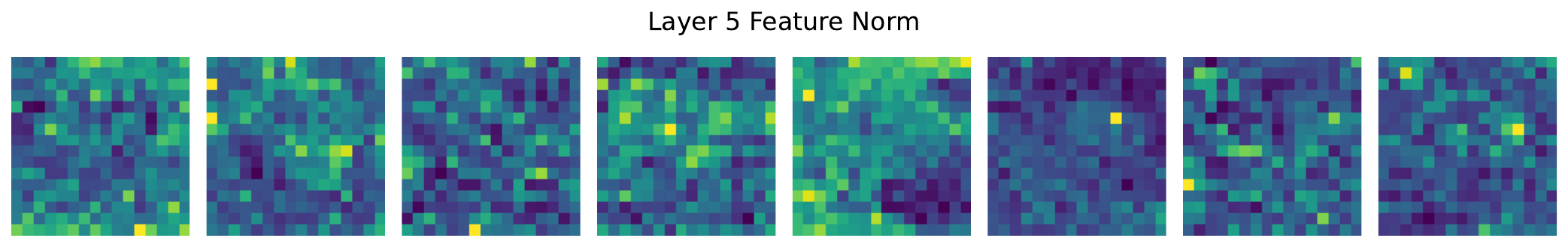}
        \label{fig:clip_5_norm}
    \end{subfigure}
    
    \label{fig:zoo_clip1}
\end{figure*}

\begin{figure*}[!hbp]
    \centering

    \begin{subfigure}{1.0\linewidth}
        \centering
        \includegraphics[width=\linewidth]{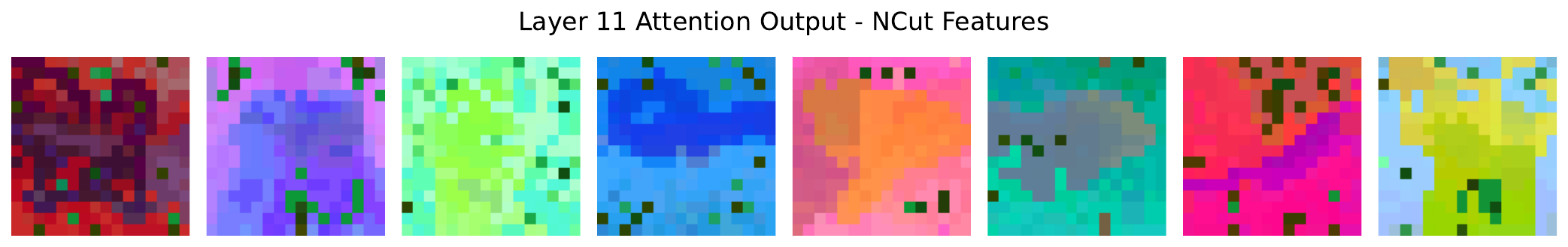}
        \label{fig:clip_11_attn}
    \end{subfigure}
    
    \vspace{-8pt}
    \begin{subfigure}{1.0\linewidth}
        \centering
        \includegraphics[width=\linewidth]{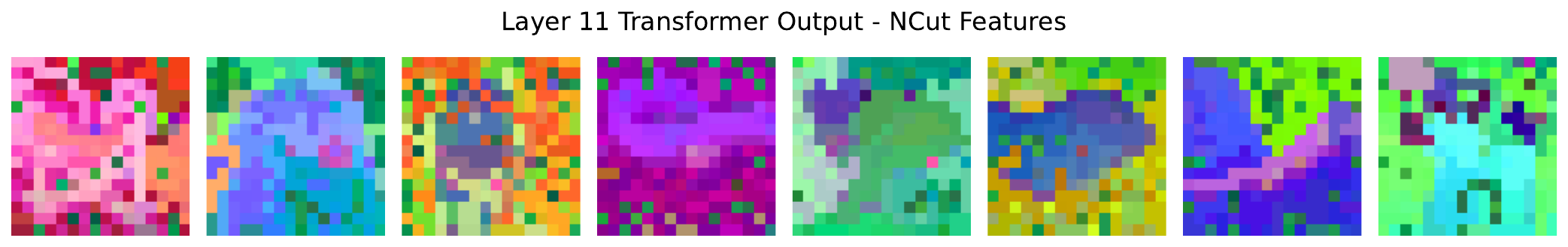}
        \label{fig:clip_11_block}
    \end{subfigure}
    
    \vspace{-8pt}
    \begin{subfigure}{1.0\linewidth}
        \centering
        \includegraphics[width=\linewidth]{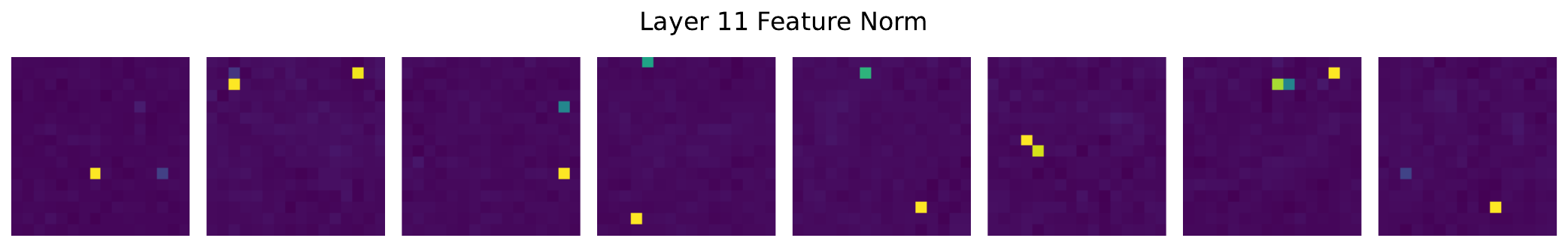}
        \label{fig:clip_11_norm}
    \end{subfigure}

    \vspace{-8pt}
    \begin{subfigure}{1.0\linewidth}
        \centering
        \includegraphics[width=\linewidth]{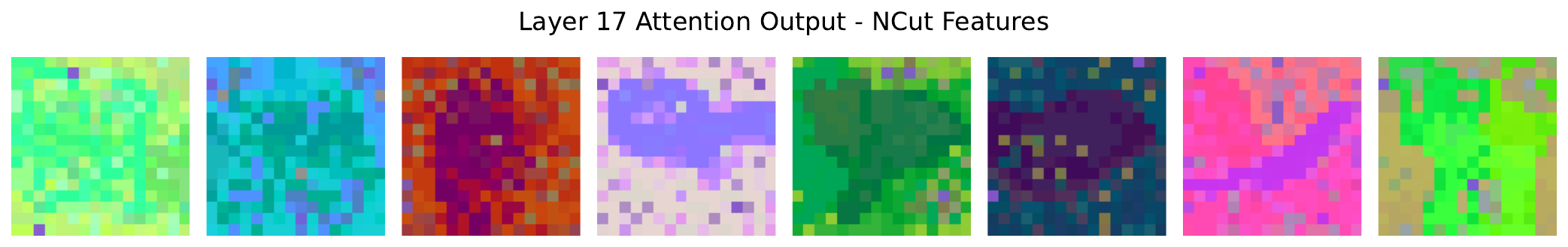}
        \label{fig:clip_17_attn}
    \end{subfigure}

    \vspace{-8pt}
    \begin{subfigure}{1.0\linewidth}
        \centering
        \includegraphics[width=\linewidth]{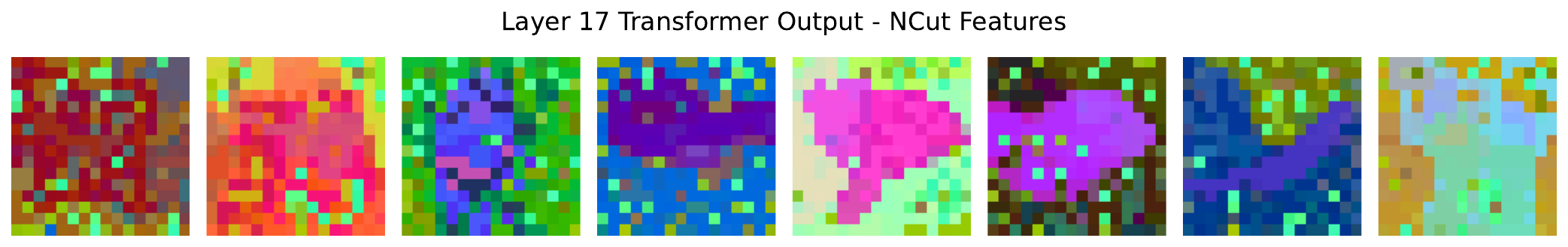}
        \label{fig:clip_17_block}
    \end{subfigure}

    \vspace{-8pt}
    \begin{subfigure}{1.0\linewidth}
        \centering
        \includegraphics[width=\linewidth]{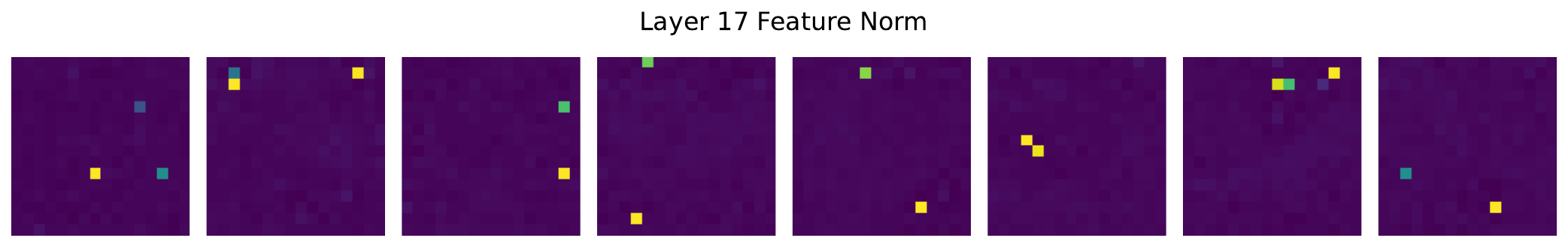}
        \label{fig:clip_17_norm}
    \end{subfigure}

    \vspace{-8pt}
    \begin{subfigure}{1.0\linewidth}
        \centering
        \includegraphics[width=\linewidth]{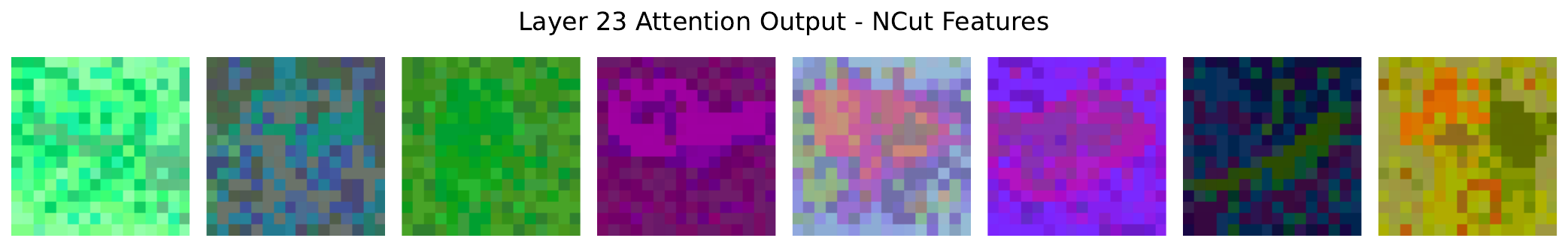}
        \label{fig:clip_23_attn}
    \end{subfigure}
    
    \label{fig:zoo_clip2}
\end{figure*}

\begin{figure*}[!htp]
    \centering

    \begin{subfigure}{1.0\linewidth}
        \centering
        \includegraphics[width=\linewidth]{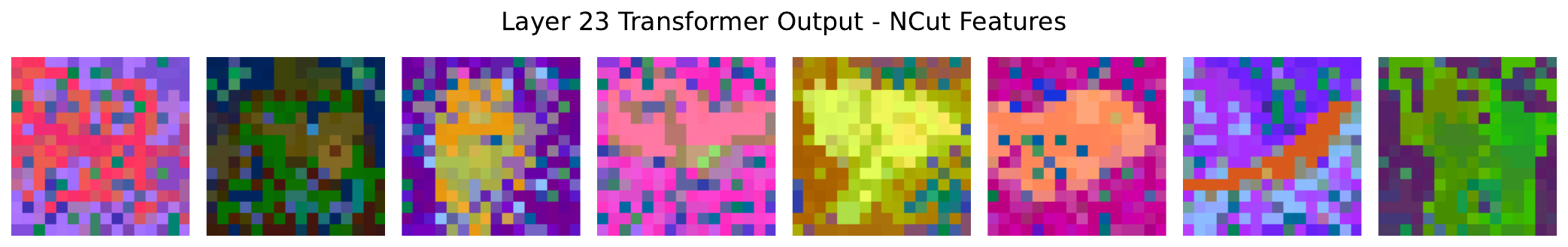}
        \label{fig:clip_23_block}
    \end{subfigure}

    \vspace{-8pt}
    \begin{subfigure}{1.0\linewidth}
        \centering
        \includegraphics[width=\linewidth]{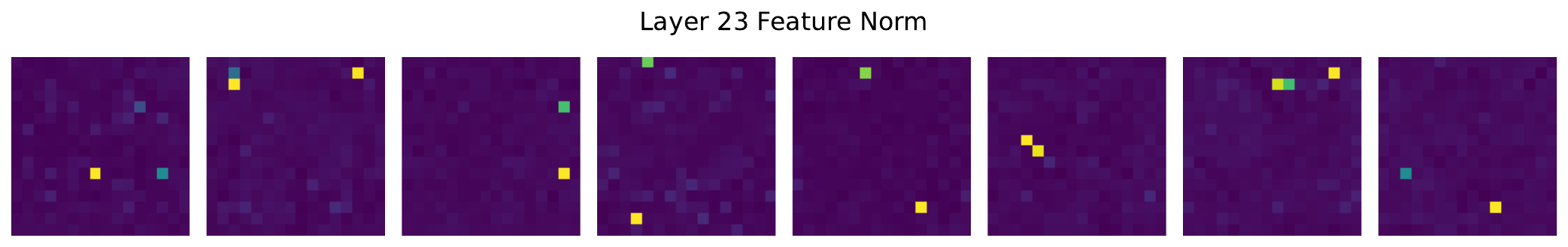}
        \label{fig:clip_23_norm}
    \end{subfigure}
    
    \label{fig:zoo_clip3}
    \vspace{128in}
\end{figure*}

\clearpage
\subsection{DINOv2 ViT L-14}

\begin{figure*}[!hbp]
    \centering
    \begin{subfigure}{1.0\linewidth}
        \centering
        \includegraphics[width=\linewidth]{figs/model_zoo_figs/input_images.pdf}
        \label{fig:dino_0_input}
    \end{subfigure}

    \vspace{-8pt}
    \begin{subfigure}{1.0\linewidth}
        \centering
        \includegraphics[width=\linewidth]{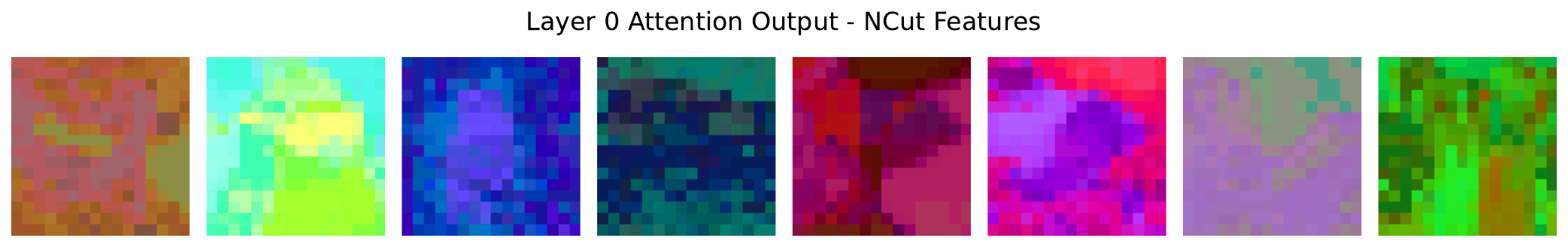}
        \label{fig:dino_0_attn}
    \end{subfigure}
    
    \vspace{-8pt}
    \begin{subfigure}{1.0\linewidth}
        \centering
        \includegraphics[width=\linewidth]{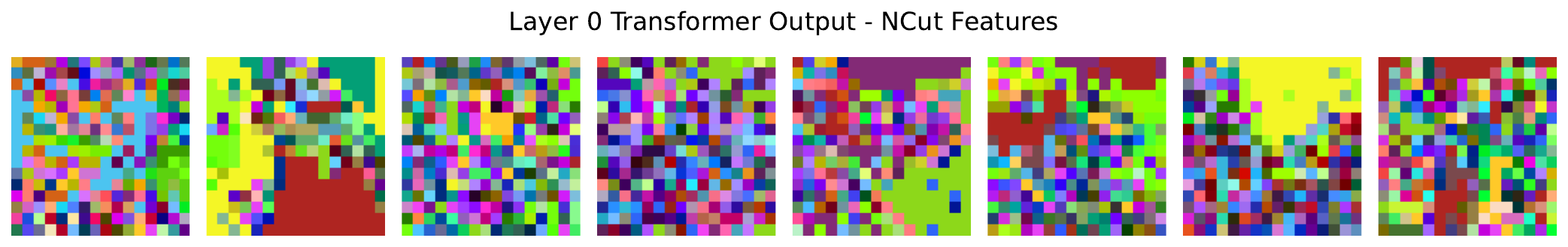}
        \label{fig:dino_0_block}
    \end{subfigure}
    
    \vspace{-8pt}
    \begin{subfigure}{1.0\linewidth}
        \centering
        \includegraphics[width=\linewidth]{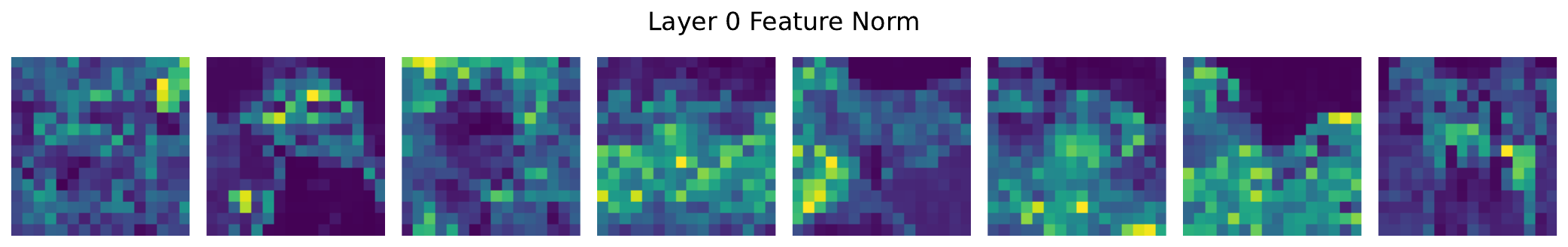}
        \label{fig:dino_0_norm}
    \end{subfigure}

    \vspace{-8pt}
    \begin{subfigure}{1.0\linewidth}
        \centering
        \includegraphics[width=\linewidth]{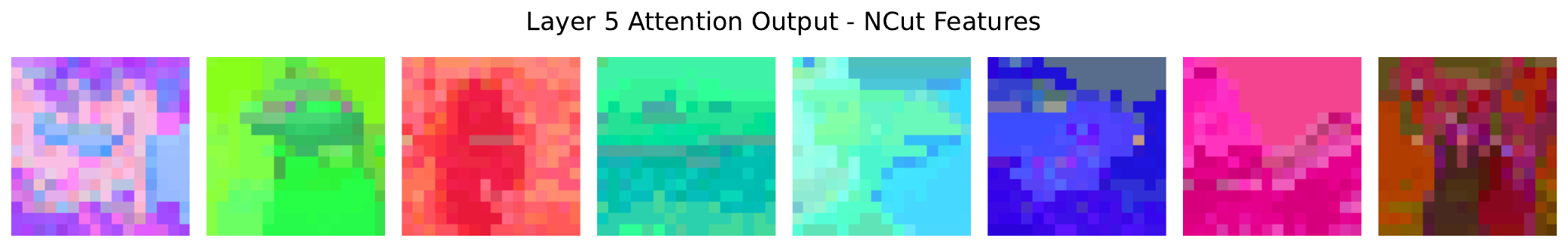}
        \label{fig:dino_5_attn}
    \end{subfigure}

    \vspace{-8pt}
    \begin{subfigure}{1.0\linewidth}
        \centering
        \includegraphics[width=\linewidth]{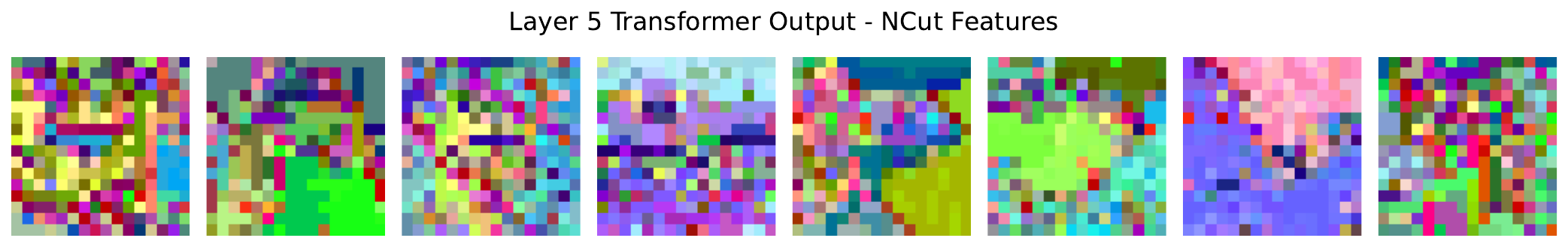}
        \label{fig:dino_5_block}
    \end{subfigure}

    \vspace{-8pt}
    \begin{subfigure}{1.0\linewidth}
        \centering
        \includegraphics[width=\linewidth]{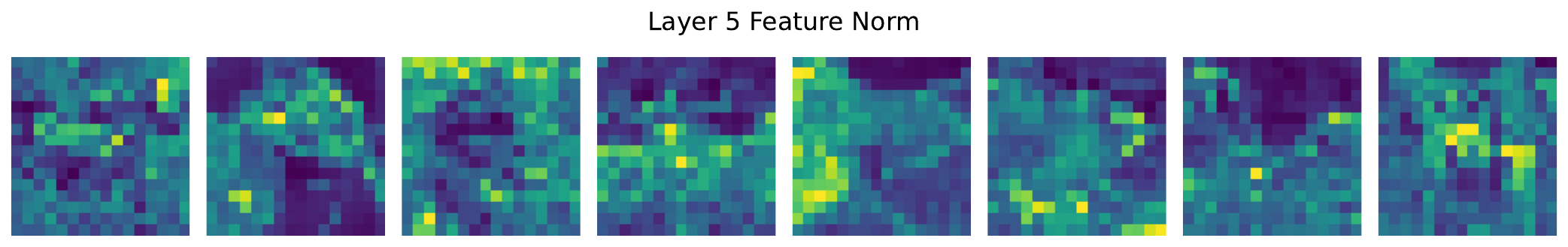}
        \label{fig:dino_5_norm}
    \end{subfigure}
    
    \label{fig:zoo_dino1}
\end{figure*}

\begin{figure*}[!hbp]
    \centering

    \begin{subfigure}{1.0\linewidth}
        \centering
        \includegraphics[width=\linewidth]{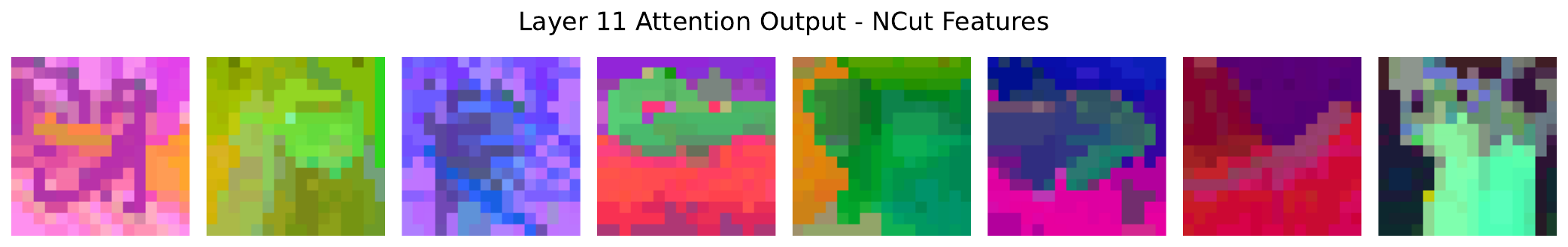}
        \label{fig:dino_11_attn}
    \end{subfigure}
    
    \vspace{-8pt}
    \begin{subfigure}{1.0\linewidth}
        \centering
        \includegraphics[width=\linewidth]{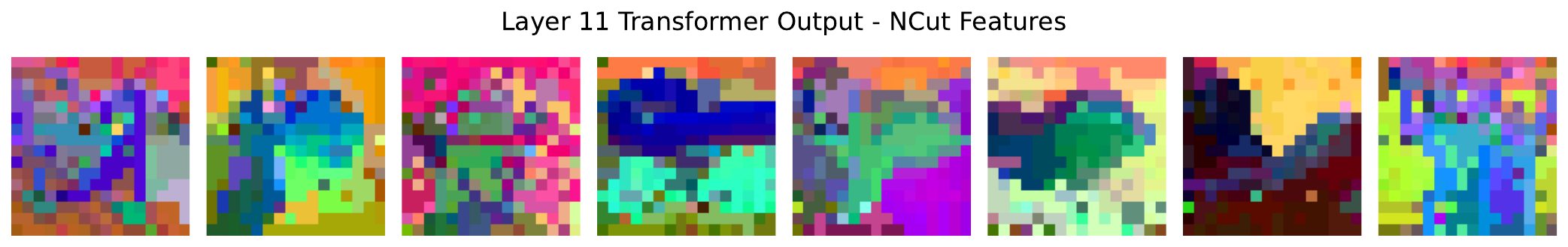}
        \label{fig:dino_11_block}
    \end{subfigure}
    
    \vspace{-8pt}
    \begin{subfigure}{1.0\linewidth}
        \centering
        \includegraphics[width=\linewidth]{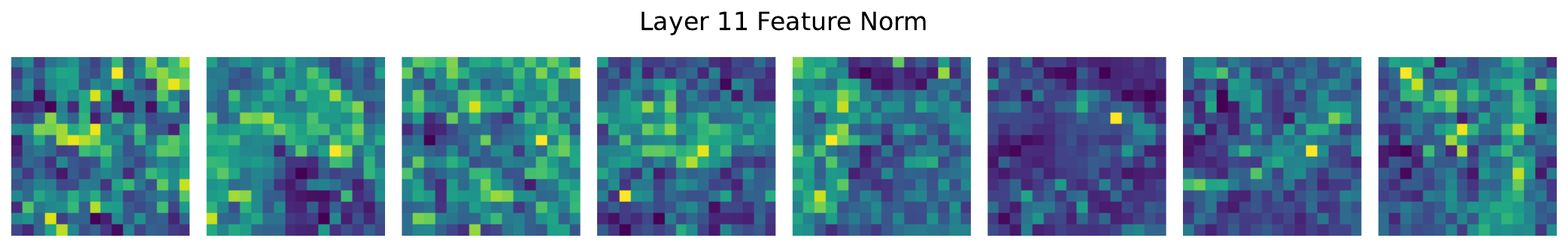}
        \label{fig:dino_11_norm}
    \end{subfigure}

    \vspace{-8pt}
    \begin{subfigure}{1.0\linewidth}
        \centering
        \includegraphics[width=\linewidth]{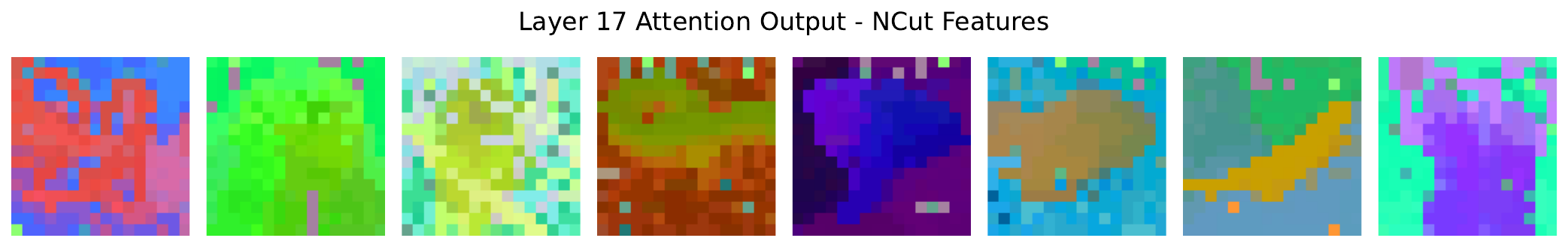}
        \label{fig:dino_17_attn}
    \end{subfigure}

    \vspace{-8pt}
    \begin{subfigure}{1.0\linewidth}
        \centering
        \includegraphics[width=\linewidth]{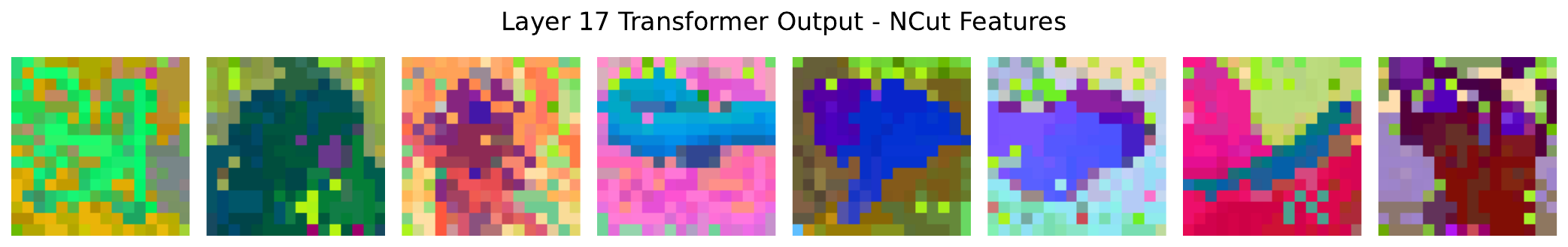}
        \label{fig:dino_17_block}
    \end{subfigure}

    \vspace{-8pt}
    \begin{subfigure}{1.0\linewidth}
        \centering
        \includegraphics[width=\linewidth]{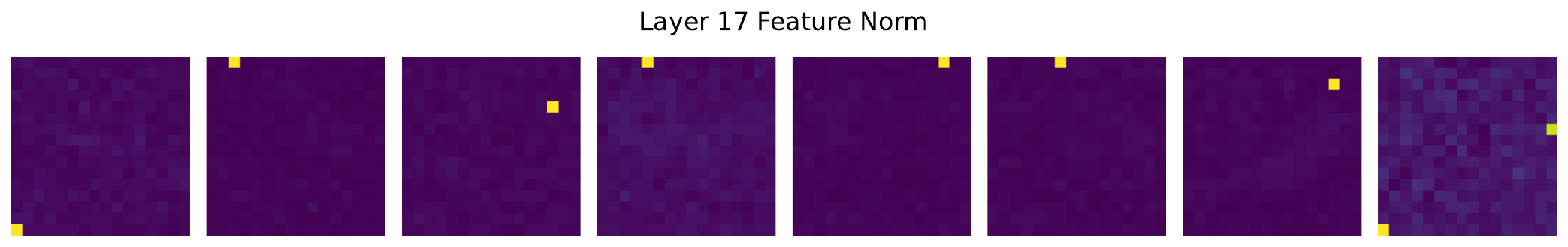}
        \label{fig:dino_17_norm}
    \end{subfigure}

    \vspace{-8pt}
    \begin{subfigure}{1.0\linewidth}
        \centering
        \includegraphics[width=\linewidth]{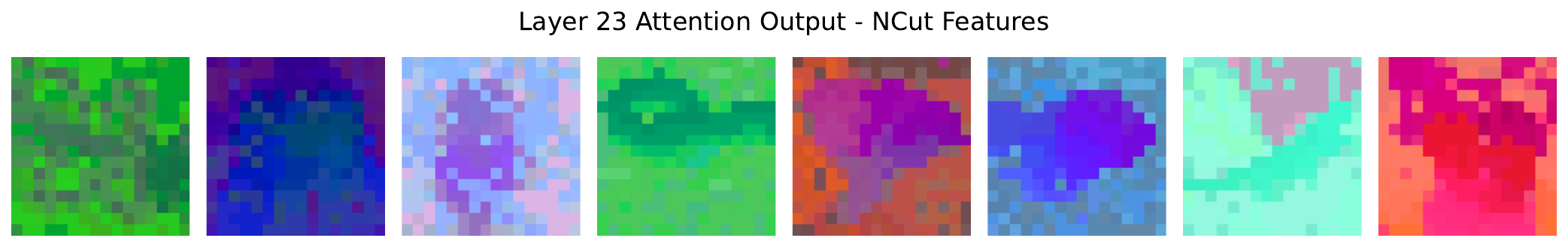}
        \label{fig:dino_23_attn}
    \end{subfigure}
    
    \label{fig:zoo_dino2}
\end{figure*}

\begin{figure*}[!htp]
    \centering

    \begin{subfigure}{1.0\linewidth}
        \centering
        \includegraphics[width=\linewidth]{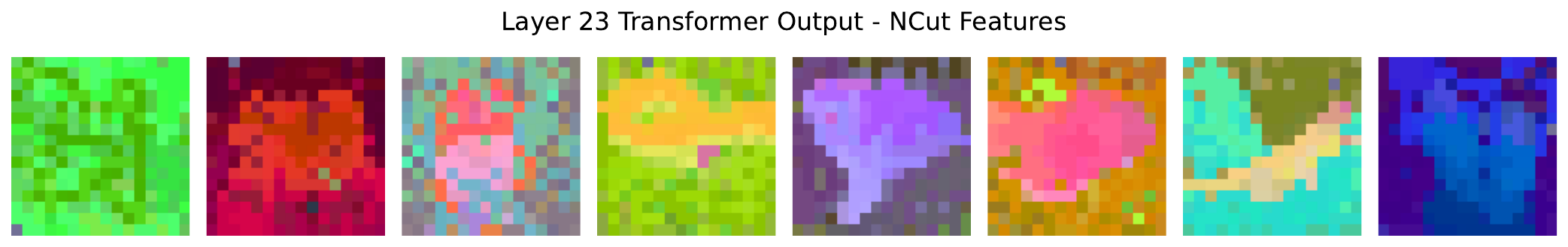}
        \label{fig:dino_23_block}
    \end{subfigure}

    \vspace{-8pt}
    \begin{subfigure}{1.0\linewidth}
        \centering
        \includegraphics[width=\linewidth]{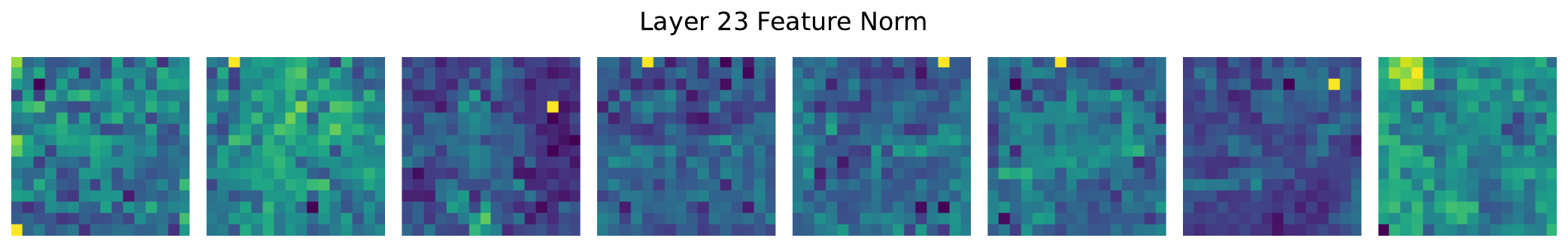}
        \label{fig:dino_23_norm}
    \end{subfigure}
    
    \label{fig:zoo_dino3}
    \vspace{128in}
\end{figure*}


\clearpage
\subsection{MAE ViT L-16 (Does not produce massive activations)}

\begin{figure*}[!hbp]
    \centering
    \begin{subfigure}{1.0\linewidth}
        \centering
        \includegraphics[width=\linewidth]{figs/model_zoo_figs/input_images.pdf}
        \label{fig:mae_0_input}
    \end{subfigure}

    \vspace{-8pt}
    \begin{subfigure}{1.0\linewidth}
        \centering
        \includegraphics[width=\linewidth]{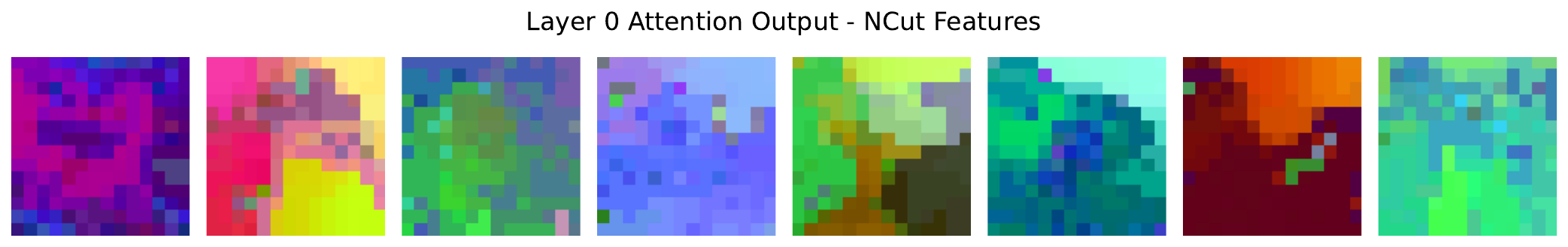}
        \label{fig:mae_0_attn}
    \end{subfigure}
    
    \vspace{-8pt}
    \begin{subfigure}{1.0\linewidth}
        \centering
        \includegraphics[width=\linewidth]{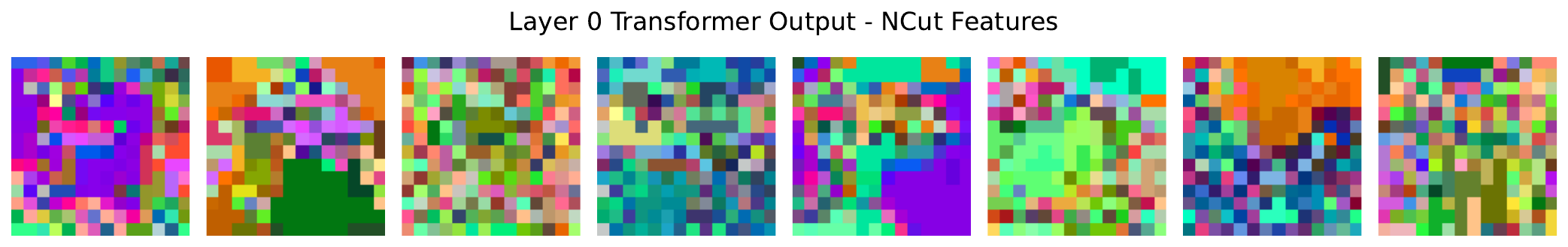}
        \label{fig:mae_0_block}
    \end{subfigure}
    
    \vspace{-8pt}
    \begin{subfigure}{1.0\linewidth}
        \centering
        \includegraphics[width=\linewidth]{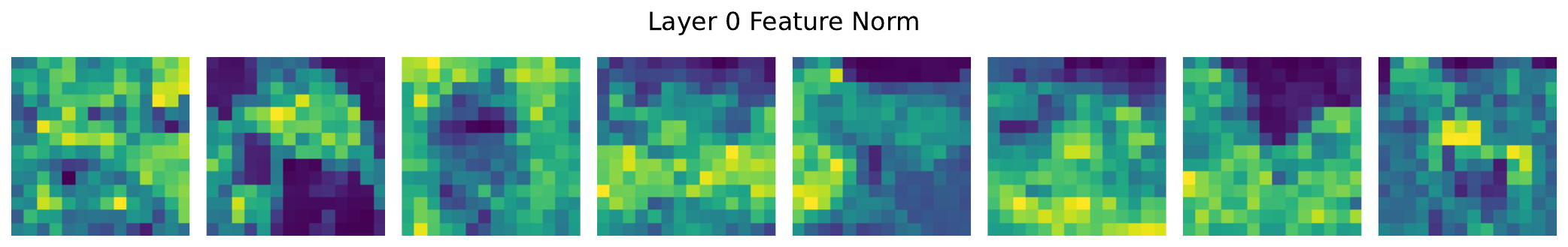}
        \label{fig:mae_0_norm}
    \end{subfigure}

    \vspace{-8pt}
    \begin{subfigure}{1.0\linewidth}
        \centering
        \includegraphics[width=\linewidth]{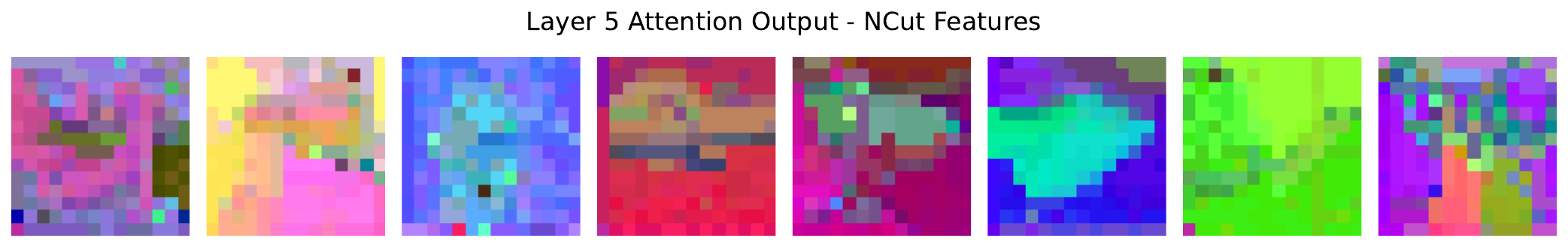}
        \label{fig:mae_5_attn}
    \end{subfigure}

    \vspace{-8pt}
    \begin{subfigure}{1.0\linewidth}
        \centering
        \includegraphics[width=\linewidth]{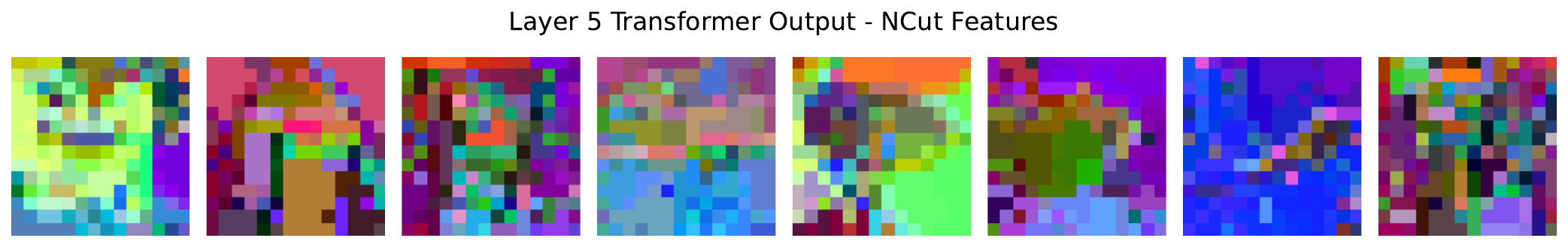}
        \label{fig:mae_5_block}
    \end{subfigure}

    \vspace{-8pt}
    \begin{subfigure}{1.0\linewidth}
        \centering
        \includegraphics[width=\linewidth]{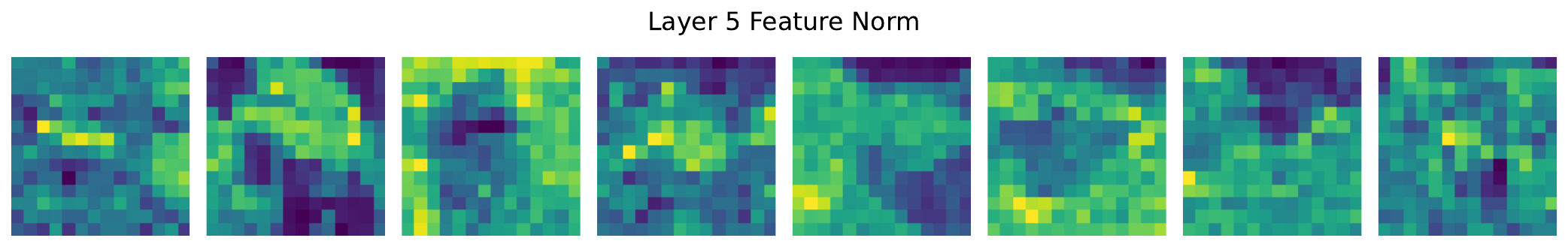}
        \label{fig:mae_5_norm}
    \end{subfigure}
    
    \label{fig:zoo_mae1}
\end{figure*}

\begin{figure*}[!hbp]
    \centering

    \begin{subfigure}{1.0\linewidth}
        \centering
        \includegraphics[width=\linewidth]{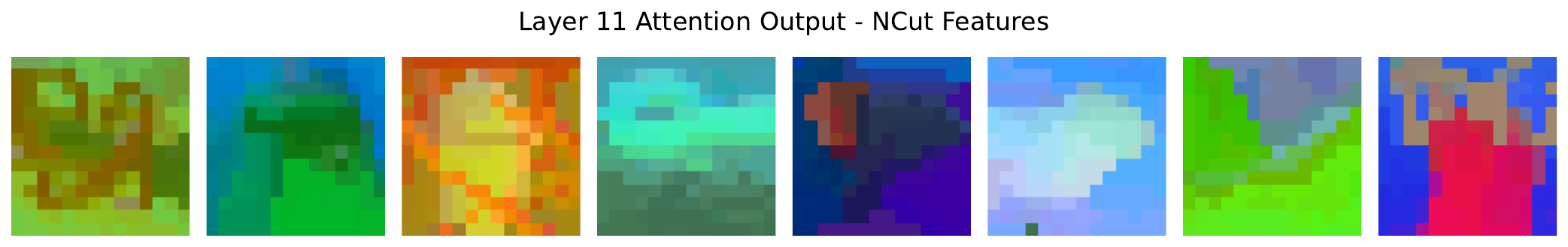}
        \label{fig:mae_11_attn}
    \end{subfigure}
    
    \vspace{-8pt}
    \begin{subfigure}{1.0\linewidth}
        \centering
        \includegraphics[width=\linewidth]{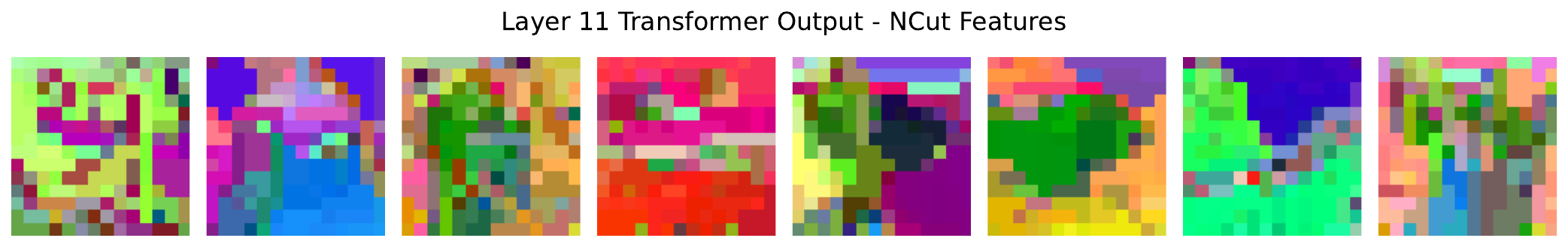}
        \label{fig:mae_11_block}
    \end{subfigure}
    
    \vspace{-8pt}
    \begin{subfigure}{1.0\linewidth}
        \centering
        \includegraphics[width=\linewidth]{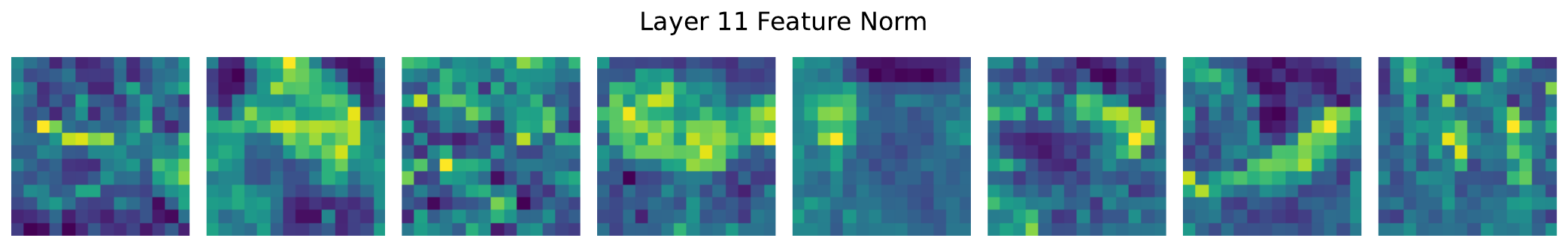}
        \label{fig:mae_11_norm}
    \end{subfigure}

    \vspace{-8pt}
    \begin{subfigure}{1.0\linewidth}
        \centering
        \includegraphics[width=\linewidth]{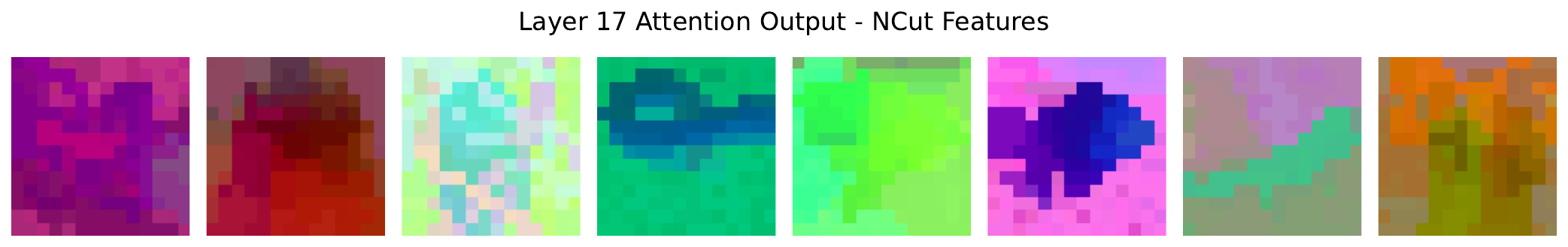}
        \label{fig:mae_17_attn}
    \end{subfigure}

    \vspace{-8pt}
    \begin{subfigure}{1.0\linewidth}
        \centering
        \includegraphics[width=\linewidth]{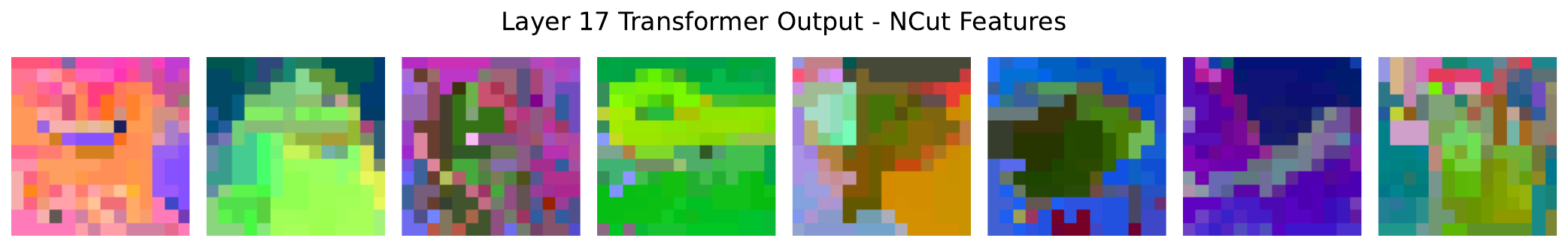}
        \label{fig:mae_17_block}
    \end{subfigure}

    \vspace{-8pt}
    \begin{subfigure}{1.0\linewidth}
        \centering
        \includegraphics[width=\linewidth]{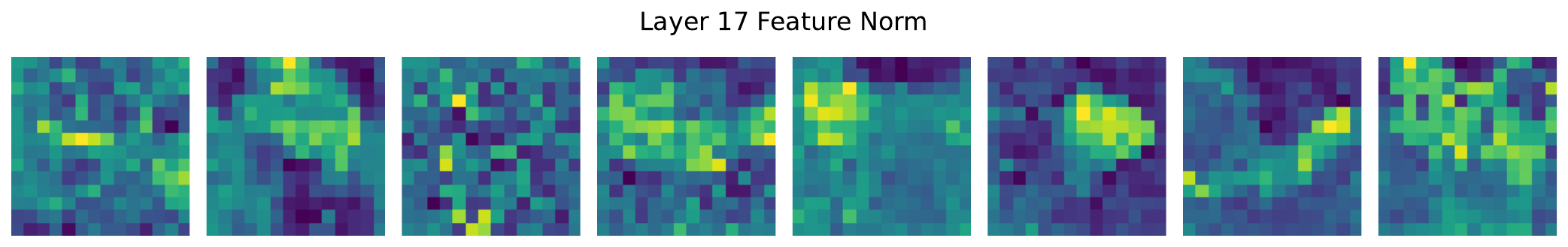}
        \label{fig:mae_17_norm}
    \end{subfigure}

    \vspace{-8pt}
    \begin{subfigure}{1.0\linewidth}
        \centering
        \includegraphics[width=\linewidth]{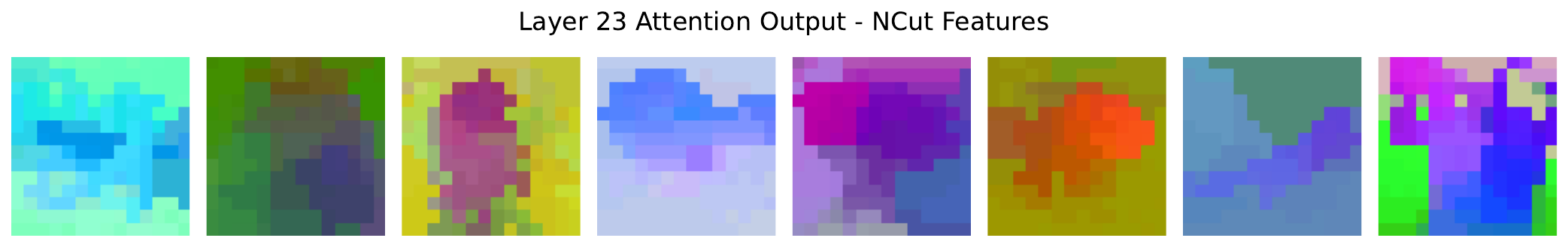}
        \label{fig:mae_23_attn}
    \end{subfigure}
    
    \label{fig:zoo_mae2}
\end{figure*}

\begin{figure*}[!htp]
    \centering
    
    \begin{subfigure}{1.0\linewidth}
        \centering
        \includegraphics[width=\linewidth]{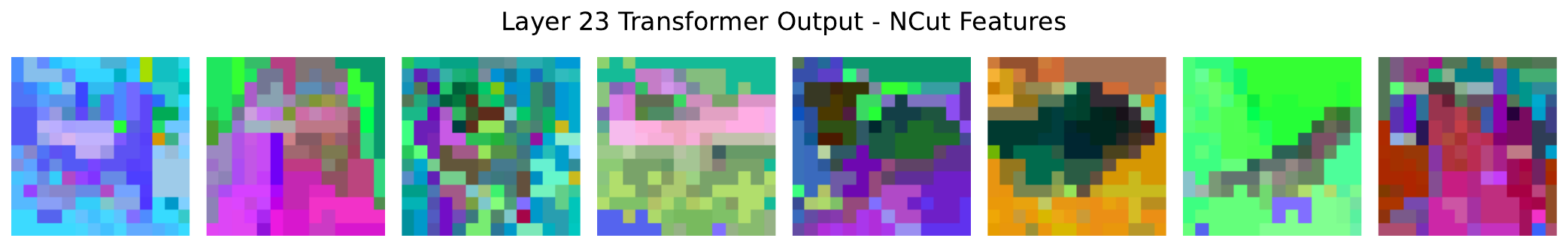}
        \label{fig:mae_23_block}
    \end{subfigure}

    \vspace{-8pt}
    \begin{subfigure}{1.0\linewidth}
        \centering
        \includegraphics[width=\linewidth]{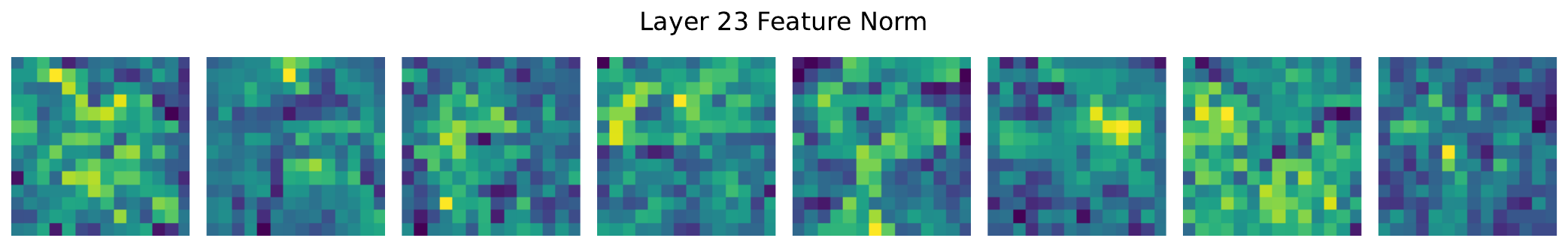}
        \label{fig:mae_23_norm}
    \end{subfigure}
    
    \label{fig:mae_dino3}
    \vspace{128in}
\end{figure*}

\end{document}